\documentclass[letterpaper]{article} 
\usepackage[preprint]{aaai2027}  
\usepackage[hyphens]{url}  
\usepackage{graphicx} 
\usepackage{natbib}  
\usepackage{caption} 
\usepackage{algorithm}
\usepackage{algorithmic}

\usepackage{newfloat}
\usepackage{listings}
\DeclareCaptionStyle{ruled}{labelfont=normalfont,labelsep=colon,strut=off} 
\floatstyle{ruled}
\newfloat{listing}{tb}{lst}{}
\floatname{listing}{Listing}

\usepackage{booktabs}

\usepackage{amsmath}
\usepackage{amsfonts}
\usepackage[inkscapelatex=false]{svg}
\usepackage{subcaption}
\usepackage{amsthm}

\newtheorem{theorem}{Theorem}

\newtheorem{lemma}{Lemma}
\newtheorem{definition}{Definition}
\newtheorem{corollary}{Corollary}

\newtheorem{remark}{Remark}

\title{ConTraIRL: Contrastive Learning of Factorized Abstractions for Robust Transferable IRL}
\author {
    Yikang Gui\textsuperscript{\rm 1,},
    Prashant Doshi\textsuperscript{\rm 1},
    Bikramjit Banerjee\textsuperscript{\rm 2}
}
\affiliations {
    \textsuperscript{\rm 1}University of Georgia\\
    \textsuperscript{\rm 2}The University of Southern Mississippi\\
    yikang.gui@uga.com, pdoshi@uga.com, Bikramjit.Banerjee@usm.edu
}

\begin{document}

\maketitle

\begin{abstract}
    Reward transfer from inverse reinforcement learning (IRL) is unreliable when behavior must generalize to unseen combinations of task dynamics and goals. We present a method for contrastive learning of factorized abstractions for transferable IRL that enables compositional reward transfer by learning decoupled latent representations of dynamics and goal. ConTraIRL uses a dual-encoder architecture to map observations into separate dynamics and goal latent spaces, optimized using two complementary contrastive objectives. Phase-conditioned contrastive learning encourages the dynamics encoder to learn goal-invariant representations, with an analogous contrastive term encouraging the goal encoder to learn dynamics-invariant representations. This factorization into dynamics and goal supports reward recovery under novel combinations of dynamics and goals. Experiments on continuous control benchmarks and a simulated robotics testbed demonstrate effective transfer to unseen dynamics-goal pairings under limited target supervision, outperforming existing transferable IRL and contrastive imitation learning methods.
    
\end{abstract}


\section{Introduction}

Inverse reinforcement learning (IRL) infers a reward function from observed expert behavior, enabling agents to acquire complex skills without manual reward design \cite{arora2021survey, gail}. Many IRL problems contain multiple sources of variation, which we refer to as factors. We focus on two types: a dynamics factor, which determines how the environment changes in response to an agent's actions, and a goal factor, which specifies the behavior the agent should achieve. Examples of dynamics factors include friction and actuator capability, while goal factors include reaching a location or maintaining a target velocity. We study compositional reward transfer, where source demonstrations cover several dynamics and goal pairings, but a target environment combines a dynamics setting and a goal that have each been observed previously, but never together~\cite{compositional-generalization,compositional-structure}. Because each expert trajectory reflects both factors, a reward model learned jointly from the observed pairings without separating their effects may encode dependencies specific to those pairings and generalize poorly to an unseen pairing~\cite{i2l,airl-me, transferability}.

We represent each environment by a context \(c=(d,g)\), where \(d\) denotes dynamics and \(g\) denotes the goal. Source contexts provide full state-only expert trajectories. In a target context \(c_{\mathrm{tar}}=(d_{\mathrm{tar}},g_{\mathrm{tar}})\), the dynamics setting and goal each appear in the source contexts but never together, and only a sparse set of expert states, called target anchors, is observed. We seek to recover a reward such that optimizing it in the target context yields a policy that, for each target factor, matches the expert behavior observed in source contexts sharing that factor.

To address this problem, we present \textbf{Con}trastive Abstractions for \textbf{Tra}nsferable \textbf{IRL} (ConTraIRL), which learns separate latent representations (abstractions) for dynamics and goals. The dynamics abstraction aligns states sharing the same dynamics across goals, while the goal abstraction aligns states sharing the same goal across dynamics. This factorization allows expert information for the two target factors to be obtained from different source contexts and combined during reward learning. The environment also provides a discrete phase label indicating each state's stage of behavioral progress. ConTraIRL uses an expert contrastive objective to structure both abstractions by factor and phase and another to separate expert and learner states at corresponding phases for reward learning.

We show that ConTraIRL induces factor-conditioned matching of expert and learner occupancies for dynamics and goals at each phase. Cross factor invariance and phase consistency (introduced in Sec.~\ref{subsec:factorized_abstractions}) allow this occupancy matching objective to transfer when each target factor appears in the source contexts. Target anchors calibrate the reward when target states differ from those observed in the source data. Experiments on four MuJoCo continuous control benchmarks and simulated assistive robotics tasks show that ConTraIRL achieves the highest normalized return across all benchmark environments and improves both return and success rate in the robotics tasks.

    \begin{figure}[!ht]
        \centering
        \includegraphics[width=0.9\linewidth]{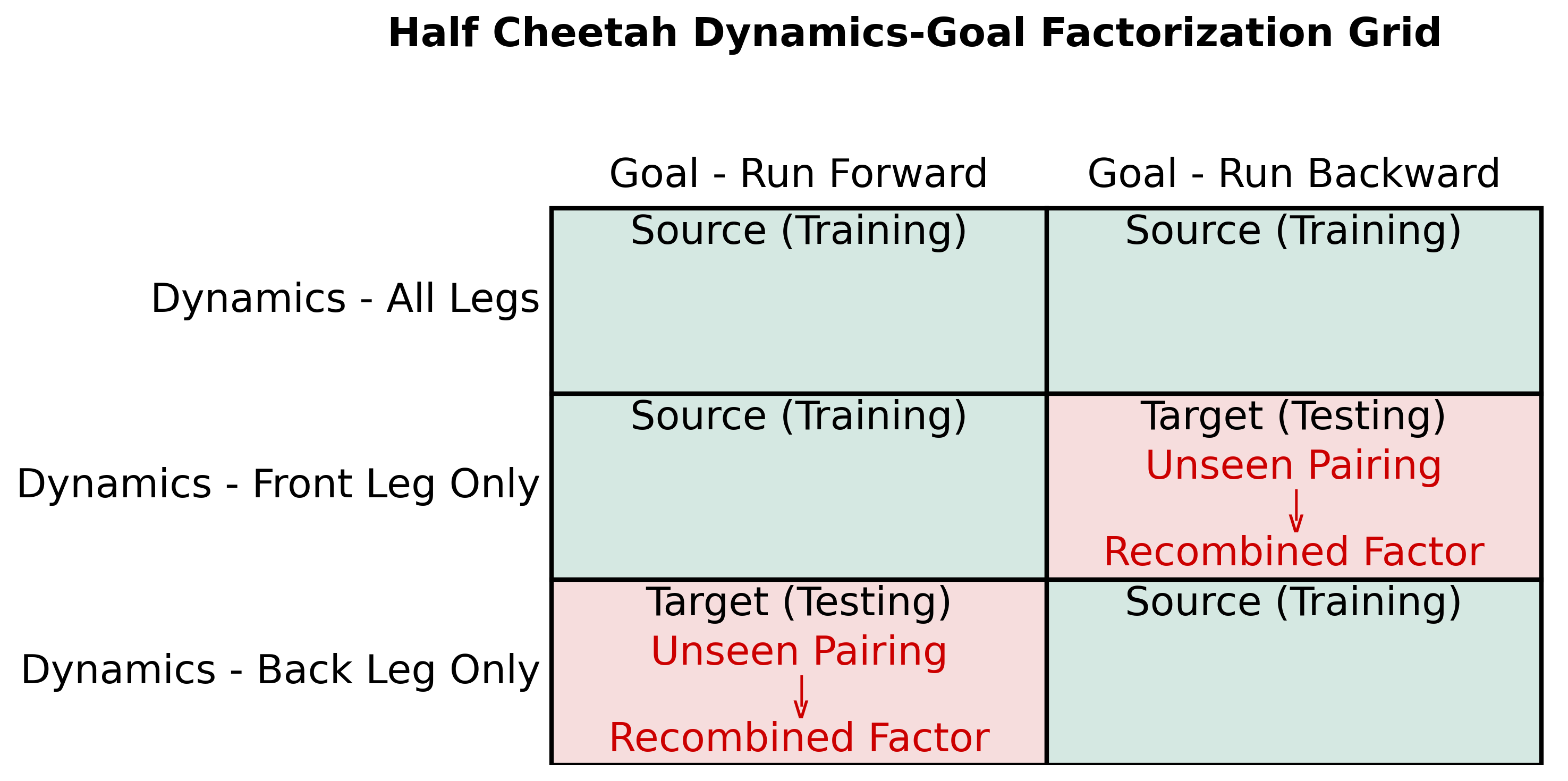}
        \caption{Illustration of factorized reward transfer. Green cells denote source contexts with complete expert trajectories, while red cells denote target contexts with limited anchors. Each factor appears in the source contexts, but its combination is unseen. ConTraIRL transfers the dynamics and goal abstractions separately to recover the target reward.}
        \label{fig:factorization}
    \end{figure}

\section{Related Work}
    \paragraph{Latent abstractions for transferable IRL.} A recent line of work learns invariant abstractions for reward transfer. XIRL~\cite{xirl} learns an embodiment invariant visual representation of task progress and defines rewards through distance to goal states in the latent space. GraphIRL~\cite{graph-irl} represents entity interactions as graphs to suppress irrelevant visual variation and learn rewards from diverse videos. TraIRL \cite{trairl} learns a shared abstract state space across multiple source tasks using a multi head VAE and recovers a reward defined over this abstraction, enabling transfer to structurally aligned target tasks without demonstrations. These methods construct a single shared abstraction that removes variation across embodiments, observations, or dynamics, but they do not separate multiple factors whose unseen combinations alter expert behavior. In contrast, ConTraIRL learns distinct dynamics and goal abstractions and recombines them for target contexts in which the individual factors are observed in source contexts but not jointly.

    \paragraph{Multiple task, intention, and transfer IRL.} Multiple task IRL methods share information across demonstrations using latent task contexts or hierarchical skills. MH-AIRL~\cite{mh-airl} learns reusable skills across tasks, while meta-IRL~\cite{meta-irl} infers a probabilistic context that conditions reward and policy adaptation. Methods for heterogeneous demonstrations instead model latent behavioral modes or multiple intentions. InfoGAIL~\cite{infogail} discovers latent factors of behavioral variation, while \(\Sigma\)-GIRL~\cite{girl} jointly cluster demonstrations and recover a reward for each intention. MSRD~\cite{msrd} separates a shared task reward from strategy specific rewards, and FLAIR~\cite{flair} extends this decomposition to lifelong adaptation from heterogeneous demonstrations. MM-ICRL~\cite{mm-icrl} infers multiple latent constraint functions from mixed demonstrations, while CoMI-IRL~\cite{comi-irl} uses contrastive learning to separate behavioral modes. These methods model heterogeneity at the level of complete tasks or behavioral modes. ConTraIRL instead separates dynamics and goal information, allowing the expert abstraction associated with each factor to be shared across contexts and composed during reward inference.

    \paragraph{Successor features.} Successor features decompose value functions into expected discounted features and reward weights, supporting transfer across tasks with shared dynamics \cite{sf,sf-gpi}. SFDE~\cite{sfde} extend this framework to environments with different dynamics by treating source successor features as noisy observations of the target successor feature function. In reward learning, \(\Psi\Phi\)-learning~\cite{psiphi} jointly estimates shared state features, agent specific successor features, and preference vectors from demonstrations, while SFM~\cite{sfm} directly matches expert and learner successor features through policy optimization and supports imitation from state only demonstrations. Successor feature based IRL methods rely on complete target trajectories to estimate the expert successor feature reference. ConTraIRL instead derives dynamics and goal expert references from source trajectories and requires only sparse target anchors.

\section{Factorized Abstractions for Compositional Reward Transfer}
    This section formalizes compositional reward transfer and develops the factorized abstraction framework underlying ConTraIRL. We first define the source-supported transfer setting. We then characterize the desired abstractions through cross-factor invariance and phase consistency and use phase-conditioned expert abstractions to construct a factorized reward. Finally, we formulate reward learning and policy optimization as a minimax objective.

    \subsection{Problem Formulation}

        We formalize our setting as a contextual Markov decision process (CMDP)~\cite{cmdp}, defined by the tuple
        \(
        (\mathcal{S},\mathcal{A},\mathcal{T},\mathcal{R},\gamma,\mathcal{C},\Psi).
        \)
        The context space is \(\mathcal{C}\subseteq\mathcal{D}\times\mathcal{G}\), where each context \(c=(d,g)\) consists of a dynamics factor \(d\in\mathcal{D}\) and a goal factor \(g\in\mathcal{G}\). The dynamics factor describes properties such as friction coefficients or joint functionality, while the goal factor describes the desired behavior, such as reaching a target location or maintaining a target velocity. 
        The transition function depends on the dynamics factor,
        \(
        \mathcal{T}(s_{t+1}\mid s_t,a_t,d),
        \)
        and \(\mathcal{R}\) is an unknown state-only reward function. The state space, action space, and discount factor are denoted by \(\mathcal{S}\), \(\mathcal{A}\), and \(\gamma\), respectively.
        
        The phase space \(\Psi=\{\psi^i\}_{i=1}^{K}\) represents discrete stages of task execution. For example, a running task may progress through standing, walking, and running phases. At each time step, the environment provides the phase label \(\psi_t\in\Psi\) associated with the current state \(s_t\), i.e. \((s_t,\psi_t)\). The phase may change as the behavior progresses. We assume that the context factors \(d\) and \(g\), together with the phase label \(\psi\), are provided by the environment. Further discussions are provided in Appendix~\ref{appendix:discussion}.
        
        Let \(\mathcal{C}_{\mathrm{src}}\subseteq\mathcal{D}\times\mathcal{G}\) denote the set of contexts for source tasks. For each \((d,g)\in\mathcal{C}_{\mathrm{src}}\), we observe a state-only expert trajectory
        \(
        \tau_E^{(d,g)}
        =
        \{(s_t^E,\psi_t^E)\}_{t=0}^{T}.
        \)
        We consider transfer to an unseen target context
        \(
        c_{\mathrm{tar}}=(d_{\mathrm{tar}},g_{\mathrm{tar}})
        \notin\mathcal{C}_{\mathrm{src}}.
        \)
        The following condition specifies the required source coverage of the target factors.
        
        \begin{definition}[Source support]
        An unseen target context
        \(c_{\mathrm{tar}}=(d_{\mathrm{tar}},g_{\mathrm{tar}})\)
        is source-supported if there exist \(g'\in\mathcal{G}\) and \(d'\in\mathcal{D}\) such that
        \(
        (d_{\mathrm{tar}},g')\in\mathcal{C}_{\mathrm{src}}
        \)
        and
        \(
        (d',g_{\mathrm{tar}})\in\mathcal{C}_{\mathrm{src}}.
        \)
        \end{definition}
        In addition to the source trajectories, we observe a small set of expert states from the target context,
        \(
        \mathcal{B}_{\mathrm{tar}}^E
        =
        \{(s_j^{\mathrm{tar}},\psi_j^{\mathrm{tar}})\}_{j=1}^{M},
        \)
        which we call \emph{anchors}. The anchors may be sparsely distributed along the target trajectory.
        
        Our objective is to learn a reward function \(\widehat{\mathcal{R}}(s,d,g, \psi)\) from the source expert trajectories and the limited target anchors such that optimizing it in an unseen target context yields a policy that matches the phase-conditioned expert behavior associated with the target dynamics and goal factors.

    \subsection{Factorized Abstractions} \label{subsec:factorized_abstractions}

        Unlike methods that encode dynamics and goal factors jointly in a single latent representation, ConTraIRL maps each state into two factor-specific latent representations. These representations lie in the latent spaces \(\mathcal{Z}_d\subseteq\mathbb{R}^{n_d}\) and \(\mathcal{Z}_g\subseteq\mathbb{R}^{n_g}\), respectively. We define factor-conditioned encoders \(\Phi_d:\mathcal{S}\times\mathcal{D} \rightarrow\mathcal{Z}_d\) and \(\Phi_g:\mathcal{S}\times\mathcal{G} \rightarrow\mathcal{Z}_g\). For a state \(s\) from context \((d,g)\) at phase \(\psi\), the corresponding abstractions are \(z_d=\Phi_d(s,d)\) and \(z_g=\Phi_g(s,g)\). The factorized abstractions of expert states should satisfy two properties: \emph{cross factor invariance} and \emph{phase consistency}.
        
        \paragraph{Cross factor invariance.}
        Cross factor invariance requires the dynamics abstraction \(z_d\) to be invariant to changes in \(g\) and the goal abstraction \(z_g\) to be invariant to changes in \(d\). At a fixed phase, states sharing the same dynamics factor should have similar dynamics abstractions across goals, while states sharing the same goal factor should have similar goal abstractions across dynamics.
        
        \paragraph{Phase consistency.}
        Phase consistency requires states sharing the same relevant factor and phase to have similar abstractions. States from different phases should remain separated, even when they share the same relevant factor.
        
        Together, consider two expert states \((s_i, \psi_i)\) and \((s_j,\psi_j)\) from contexts \(c_i=(d_i,g_i)\) and \(c_j=(d_j,g_j)\). Cross factor invariance  and phase consistency are expressed as
        \begin{equation}
        \begin{aligned}
        d_i=d_j,\psi_i=\psi_j
        &\Longleftrightarrow
        \Phi_d(s_i,d_i)\approx\Phi_d(s_j,d_j),\\
        g_i=g_j,\psi_i=\psi_j
        &\Longleftrightarrow
        \Phi_g(s_i,g_i)\approx\Phi_g(s_j,g_j).
        \end{aligned}
        \label{eq:abstraction_properties}
        \end{equation}
        These properties yield factorized abstractions that capture the respective factor and behavioral phase while remaining invariant to the other factor. This factorization enables dynamics and goal information to transfer independently to unseen combinations as illustrated in Sec.~\ref{sec:theoretical_analysis}.

    \subsection{Reward Learning from Abstractions} \label{sec:reward_learning}
        We next construct a reward from the factorized abstractions. For each factor and phase, we summarize the corresponding source expert abstractions by their expectation. The reward for a learner state is defined as the sum of its dot products with the dynamics and goal expert abstractions at the same phase. Because these abstractions are parameterized by the encoders, reward learning and policy optimization form a minimax problem: the encoders separate expert and learner occupancies, while the policy reduces this separation.

        \paragraph{Phase-conditioned expert abstractions.} Let \(\rho_E(\cdot\mid d,\psi)\) and \(\rho_E(\cdot\mid g,\psi)\) denote the source expert state occupancies conditioned on dynamics \(d\) and goal \(g\), respectively, at phase \(\psi\). For fixed factor values \(d\) and \(g\), the phase-conditioned expert abstractions are defined as
        \begin{equation}
        \begin{aligned}
        z_d^E(\psi)
        &=
        \mathbb{E}_{s\sim\rho_E(\cdot\mid d,\psi)}
        \left[\Phi_d(s,d)\right],\\
        z_g^E(\psi)
        &=
        \mathbb{E}_{s\sim\rho_E(\cdot\mid g,\psi)}
        \left[\Phi_g(s,g)\right].
        \end{aligned}
        \end{equation}
        For a source-supported target context, the dynamics abstraction is obtained from source trajectories containing \(d_{\mathrm{tar}}\), while the goal abstraction is obtained from source trajectories containing \(g_{\mathrm{tar}}\). 

        \paragraph{Reward function.} We constrain the encoder outputs to lie in their respective unit balls: \(\|\Phi_d(s,d)\|_2\leq 1\) and \(\|\Phi_g(s,g)\|_2\leq 1\). Given a learner state \(s\) from context \((d,g)\) at phase \(\psi\), we define the reward as
        \begin{equation}
        \widehat{\mathcal{R}}(s,d,g,\psi)
        =
        \Phi_d(s,d)^\top z_d^E(\psi)
        +
        \Phi_g(s,g)^\top z_g^E(\psi).
        \label{eq:factorized_reward}
        \end{equation}
        The first term measures the similarity between the learner and expert dynamics abstractions, while the second measures the similarity between their goal abstractions at the same phase.

        \paragraph{Reward learning criterion.}
        Recall that apprenticeship learning~\cite{al} optimizes
        \begin{equation}
        \min_{\pi}\max_{r}
        \mathbb{E}_{s\sim\rho_E}[r(s)]
        -
        \mathbb{E}_{s\sim\rho_\pi}[r(s)].
        \label{eq:al_objective}
        \end{equation}        
        Under the factorized reward parameterization in Eq.~\eqref{eq:factorized_reward}, reward maximization is performed over the encoders \(\Phi_d\) and \(\Phi_g\). Let \(\rho_E\) and \(\rho_\pi\) denote the expert and learner occupancies over \((s,d,g,\psi)\). Substituting the factorized reward into Eq.~\eqref{eq:al_objective} gives
        \begin{equation}
        \min_{\pi}\max_{\Phi_d,\Phi_g}
        \mathbb{E}_{\rho_E}
        \left[\widehat{\mathcal{R}}(s,d,g,\psi)\right]
        -
        \mathbb{E}_{\rho_\pi}
        \left[\widehat{\mathcal{R}}(s,d,g,\psi)\right].
        \label{eq:contrairl_al_objective}
        \end{equation}
        The encoders maximize the expected return gap between expert and learner occupancies, while the policy minimizes this gap by increasing the learner return.

\section{ConTraIRL}
    We now present the ConTraIRL algorithm. We first describe the factor-modulated dynamics and goal encoders. We then introduce two contrastive objectives: the first enforces cross factor invariance and phase consistency using expert trajectories, while the second separates expert and learner states to optimize the reward. Finally, we provide a theoretical analysis of the learned reward and its compositional transfer.

    \begin{figure*}[t]
    \centering
    \begin{subfigure}[t]{0.62\textwidth}
        \centering
        \includegraphics[width=\linewidth]{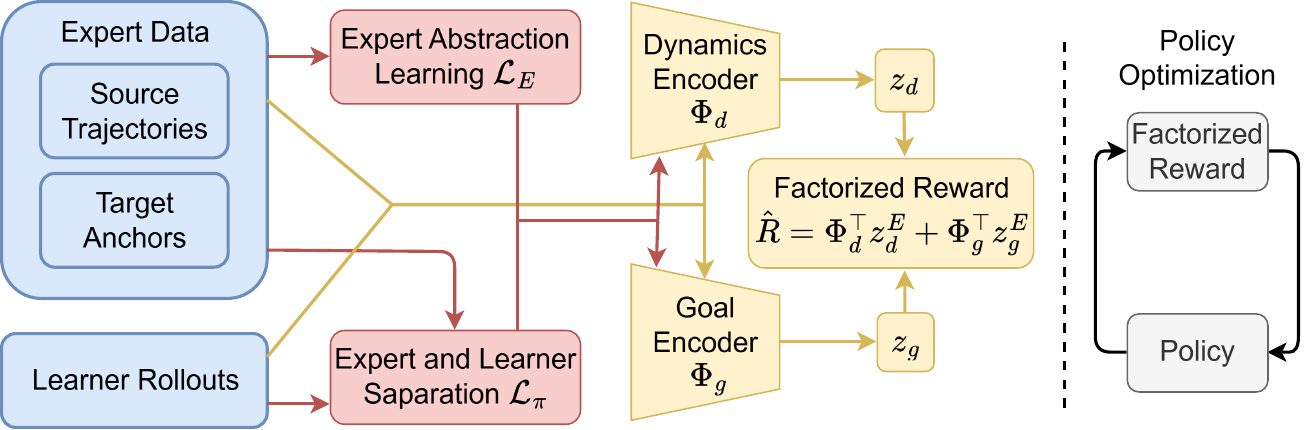}
        \caption{}
        \label{fig:contrairl_overview}
    \end{subfigure} 
    \hfill
    \begin{subfigure}[t]{0.30\textwidth}
        \centering
        \includegraphics[width=\linewidth]{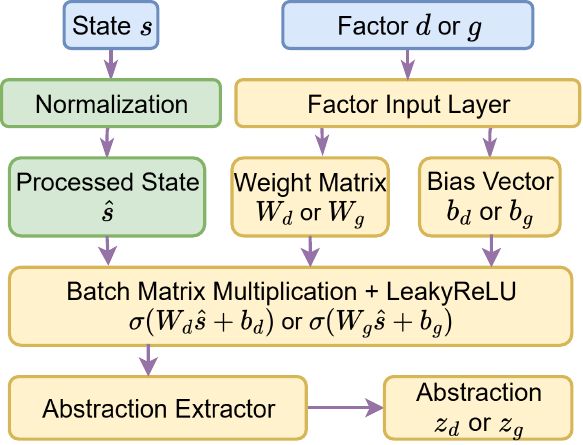}
        \caption{}
        \label{fig:context_encoder}
    \end{subfigure}
    \caption{Architecture of ConTraIRL. (a) ConTraIRL learns dynamics and goal abstractions and combines them to construct the reward for the unseen target context. (b) The Factor-Modulated Encoder maps states into factor-specific latent spaces.}
    \label{fig:contrairl_architecture}
\end{figure*}

    \subsection{Factor-Modulated Encoder}\label{subsec:encoder_arch}

        ConTraIRL implements the dynamics encoder \(\Phi_d\) and goal encoder \(\Phi_g\) as separate networks. Rather than concatenating the corresponding factor with the state, each encoder uses an input layer inspired by hypernetworks \citep{hypernetworks}. A small multilayer perceptron generates the parameters of an affine state transformation from the factor value:
        \begin{equation}
        h_d=\sigma\!\left(W_d(d)s+b_d(d)\right),        
        h_g=\sigma\!\left(W_g(g)s+b_g(g)\right),
        \end{equation}
        where \(\sigma\) is a nonlinear activation function.
        These transformations allow the dynamics and goal factors to modulate state feature extraction separately. The resulting features are passed through abstraction extractor to produce \(z_d\) and \(z_g\), as illustrated in Figure~\ref{fig:context_encoder}.

    \subsection{Contrastive Learning of Abstractions}

        We train the encoders using two contrastive objectives. The first enforces the abstraction properties in Eq.~\eqref{eq:abstraction_properties} using pairs of expert states. The second contrasts expert and learner states to implement the inner maximization in Eq.~\eqref{eq:contrairl_al_objective}.
        
        \paragraph{Expert abstraction learning (\(\mathcal{L}_{E}\)).}
        Consider two expert states \((s_i,\psi_i)\) and \((s_j,\psi_j)\) with context factors \((d_i,g_i)\) and \((d_j,g_j)\). We set \(y_{ij}^{d}=1\) if \(d_i=d_j\) and \(\psi_i=\psi_j\), and \(y_{ij}^{g}=1\) if \(g_i=g_j\) and \(\psi_i=\psi_j\); both labels are \(-1\) otherwise. To enforce invariance to the other factor, positive samples include dynamics pairs with \(g_i\neq g_j\) and goal pairs with \(d_i\neq d_j\).        
        For two abstractions \(z_i\) and \(z_j\), we define
        \begin{equation}
        \ell(z_i,z_j,y)=
        \begin{cases}
        1-z_i^\top z_j, & y=1,\\
        \max\left(0,z_i^\top z_j-\epsilon\right), & y=-1,
        \end{cases}
        \label{eq:expert_pair_loss}
        \end{equation}
        where \(\epsilon\) is the similarity margin for negative pairs. Let \(z_{d,k}^{E}=\Phi_d(s_k^{E},d_k^{E})\) and \(z_{g,k}^{E}=\Phi_g(s_k^{E},g_k^{E})\) denote the expert abstractions for \(k\in\{i,j\}\). The expert abstraction objective is
        \begin{equation}
            \mathcal{L}_{E}
            =
            \mathbb{E}_{(i,j)\sim\mathcal{D}_{E}}
            \left[
            \ell(z_{d,i}^{E},z_{d,j}^{E},y_{ij}^{d})
            +
            \ell(z_{g,i}^{E},z_{g,j}^{E},y_{ij}^{g})
            \right],
            \label{eq:expert_abstraction_loss}
        \end{equation}
        where \(\mathcal{D}_{E}\) denotes the distribution of expert state pairs.

        \paragraph{Expert and learner separation (\(\mathcal{L}_{\pi}\)).}
        For each encoder, we sample two expert states \((s_i^{E},\psi_i^{E})\) and \((s_j^{E},\psi_j^{E})\), together with learner states \(\{(s_k^\pi,\psi_k^\pi)\}_{k=1}^{N}\). For \(\Phi_d\), the samples satisfy \(d_i^{E}=d_j^{E}=d_k^\pi\) and \(\psi_i^{E}=\psi_j^{E}=\psi_k^\pi\) for every \(k\); for \(\Phi_g\), they satisfy \(g_i^{E}=g_j^{E}=g_k^\pi\) and \(\psi_i^{E}=\psi_j^{E}=\psi_k^\pi\). We define the learner abstractions as \(z_{d,k}^\pi=\Phi_d(s_k^\pi,d_k^\pi)\) and \(z_{g,k}^\pi=\Phi_g(s_k^\pi,g_k^\pi)\).        
        The objective of expert and learner separation follows InfoNCE~\cite{infonce}:
        \begin{equation}
        \begin{aligned}
        \mathcal{L}_{\pi}
        =
        -\mathbb{E}\Bigg[
        &\log
        \frac{
        \exp(\hat z_{d,i}^{\top}\hat z_{d,j}/\tau)
        }{
        \exp(\hat z_{d,i}^{\top}\hat z_{d,j}/\tau)
        +
        \sum_{k=1}^{N}
        \exp(\hat z_{d,i}^{\top}z_{d,k}^{\pi}/\tau)
        }\Bigg]\\
        -\mathbb{E}\Bigg[
        &\log
        \frac{
        \exp(\hat z_{g,i}^{\top}\hat z_{g,j}/\tau)
        }{
        \exp(\hat z_{g,i}^{\top}\hat z_{g,j}/\tau)
        +
        \sum_{k=1}^{N}
        \exp(\hat z_{g,i}^{\top}z_{g,k}^{\pi}/\tau)
        }
        \Bigg].
        \end{aligned}
        \label{eq:expert_learner_loss}
        \end{equation}
        
        The encoders minimize \(\mathcal{L}=\mathcal{L}_{E}+\mathcal{L}_{\pi}\). The full algorithm is provided in Appendix~\ref{appendix:algorithm}.

    \subsection{Theoretical Analysis} \label{sec:theoretical_analysis}

        We show that, under the first-order approximation, \(\mathcal L_\pi\) implements phase-conditioned occupancy matching and that, under the abstraction properties targeted by \(\mathcal L_E\), this matching transfers to source supported target contexts.
        
        Let \(\rho_E(\cdot\mid d,\psi)\) denote the expert state occupancy conditioned on dynamics factor \(d\) and phase \(\psi\), and let \(\rho_E(\cdot\mid g,\psi)\) denote the expert state occupancy conditioned on goal factor \(g\) and phase \(\psi\). Similarly, let \(\rho_\pi(\cdot\mid d,\psi)\) and \(\rho_\pi(\cdot\mid g,\psi)\) denote the corresponding learner state occupancies induced by policy \(\pi\). For fixed \(d\), \(g\), and \(\psi\), define
        \begin{equation}
        \begin{aligned}
        \Delta_d(\Phi_d,\pi)
        &=
        \mathbb E_{\rho_E(\cdot\mid d,\psi)}
        \left[\Phi_d(s,d)^\top z_d^E(\psi)\right]\\
        &\phantom{=}
        -
        \mathbb E_{\rho_\pi(\cdot\mid d,\psi)}
        \left[\Phi_d(s,d)^\top z_d^E(\psi)\right],\\
        \Delta_g(\Phi_g,\pi)
        &=
        \mathbb E_{\rho_E(\cdot\mid g,\psi)}
        \left[\Phi_g(s,g)^\top z_g^E(\psi)\right]\\
        &\phantom{=}
        -
        \mathbb E_{\rho_\pi(\cdot\mid g,\psi)}
        \left[\Phi_g(s,g)^\top z_g^E(\psi)\right].
        \end{aligned}
        \nonumber
        \end{equation}

        \begin{lemma} \label{lemma:objective_equivalence}
        Under a first-order Taylor approximation of the log-sum-exp terms in \(\mathcal L_\pi\), minimizing the resulting objective with respect to \(\Phi_d\) and \(\Phi_g\) is equivalent to the inner reward maximization in Eq.~\eqref{eq:contrairl_al_objective}, namely, maximizing the factorized expert-learner reward gap \(\Delta_d(\Phi_d,\pi)+\Delta_g(\Phi_g,\pi)\).
        \end{lemma}
        
        To relate the ConTraIRL objective to occupancy matching, we define the factor-conditioned contrastive discrepancy over this class as
        \begin{equation}
        \mathcal D_{\mathrm{ConTra}}(\rho_E,\rho_\pi)
        =
        \max_{\Phi_d,\Phi_g}
        \left[
        \Delta_d(\Phi_d,\pi)
        +
        \Delta_g(\Phi_g,\pi)
        \right].
        \nonumber
        \end{equation}
        Let \(D_{\mathrm{TV}}\) denote the total variation distance. We define
        \begin{align}
        D_{\mathrm{TV}}^d
        &=
        D_{\mathrm{TV}}\left(
        \rho_E(\cdot\mid d,\psi),
        \rho_\pi(\cdot\mid d,\psi)
        \right), \nonumber\\
        D_{\mathrm{TV}}^g
        &=
        D_{\mathrm{TV}}\left(
        \rho_E(\cdot\mid g,\psi),
        \rho_\pi(\cdot\mid g,\psi)
        \right). \nonumber
        \end{align}
        
        \begin{theorem}[Factor-conditioned occupancy matching] \label{theorem:occupancy_matching}
        The factor-conditioned contrastive discrepancy satisfies
        \begin{equation}
        \frac{1}{4}
        \left(D_{\mathrm{TV}}^d+D_{\mathrm{TV}}^g\right)
        \leq
        \mathcal D_{\mathrm{ConTra}}(\rho_E,\rho_\pi)
        \leq
        2
        \left(D_{\mathrm{TV}}^d+D_{\mathrm{TV}}^g\right).
        \nonumber
        \end{equation}
        \end{theorem}
        Proofs are provided in Appendix~\ref{appendix:proof}. Theorem~\ref{theorem:occupancy_matching} interprets encoder maximization as dual reward optimization and policy optimization as minimizing the resulting occupancy discrepancy. GAIL \citep{gail} induces the Jensen Shannon divergence, whereas ConTraIRL induces a discrepancy equivalent to the sum of the phase conditioned dynamics and goal total variation distances.
        For a source supported target context with unknown expert occupancy, cross factor invariance and phase consistency allow the source expert abstractions for \(d_{\mathrm{tar}}\) and \(g_{\mathrm{tar}}\) to define the reward for their unseen combination.

        \begin{corollary} \label{corollary:compositional_transfer}
            Let \(c_{\mathrm{tar}}=(d_{\mathrm{tar}},g_{\mathrm{tar}})\) be source-supported, and suppose that cross-factor invariance and phase consistency hold exactly and that there exists a target policy simultaneously matches \(z_d^E(\psi)\) and \(z_g^E(\psi)\) in expectation for every \(\psi\in\Psi\). Then, at a minimax solution \((\widehat{\Phi}_d,\widehat{\Phi}_g, \widehat{\pi}_{\mathrm{tar}})\), for every \(\psi\in\Psi\),
            \begin{equation}
                \begin{aligned}
                \mathbb{E}_{s\sim
                \rho_{\widehat{\pi}_{\mathrm{tar}}}
                (\cdot\mid c_{\mathrm{tar}},\psi)}
                \left[\widehat{\Phi}_d(s,d_{\mathrm{tar}})\right]
                &=
                z_d^E(\psi),\\
                \mathbb{E}_{s\sim
                \rho_{\widehat{\pi}_{\mathrm{tar}}}
                (\cdot\mid c_{\mathrm{tar}},\psi)}
                \left[\widehat{\Phi}_g(s,g_{\mathrm{tar}})\right]
                &=
                z_g^E(\psi),
                \end{aligned}
                \nonumber
            \end{equation}
            where \(z_d^E(\psi)\) and \(z_g^E(\psi)\) are evaluated using \(\widehat{\Phi}_d\) and \(\widehat{\Phi}_g\), respectively. Thus, the transferred target policy matches the source expert expected abstractions associated with each target factor and phase.
        \end{corollary}
        
        When a target factor-conditioned occupancy differs from its source counterpart, target states associated with the same factor and phase may not be represented by the source trajectories. The target anchors provide labeled examples of these shifted states. For an anchor \((s_j^{\mathrm{tar}},\psi_j^{\mathrm{tar}})\), we pair it in \(\mathcal L_E\) with source expert states sharing \(d_{\mathrm{tar}}\) and \(\psi_j^{\mathrm{tar}}\) for \(\Phi_d\), and with source expert states sharing \(g_{\mathrm{tar}}\) and \(\psi_j^{\mathrm{tar}}\) for \(\Phi_g\).
        
        \begin{remark} \label{claim:target_anchor}
        When the source and target occupancies differ, target anchors align target states with source-derived abstractions for the same factor and phase. This alignment calibrates the transferred reward at the observed target states and facilitates factor-conditioned occupancy matching.
        \end{remark}
        Corollary~\ref{corollary:compositional_transfer} and Remark~\ref{claim:target_anchor} characterize source supported transfer and target specific correction, respectively. Source trajectories provide the expert behavior associated with each target factor, while target anchors account for occupancy shifts that cannot be identified from source contexts alone.

\section{Experiments} \label{sec:experiments}

    We evaluate whether the learned reward transfers to unseen combinations of dynamics and goal factors. Source contexts provide full expert trajectories, while target contexts provide only a small set of expert states with phase labels.
    ConTraIRL is trained jointly across all contexts. Unless otherwise specified, \(20\%\) of the states from one target expert trajectory are randomly sampled as anchors at the beginning of training and remain fixed throughout training. No other target expert states are accessible.
    We optimize a policy under the learned reward and report its return under the true environment reward. The environment provides a phase \(\psi\) for each \(s\). 

    \paragraph{Benchmarks.}
    We evaluate ConTraIRL on four MuJoCo continuous control environments: \texttt{Ant}, \texttt{HalfCheetah}, \texttt{Walker2d}, and \texttt{Swimmer}. For each environment, we independently vary the dynamics and goal factors to construct multiple context combinations. Changes in dynamics produce different expert trajectories and occupancies even for a fixed goal. Jointly varying dynamics and goals therefore requires IRL to separate goal-related behavior from dynamics-induced variation, creating a challenging setting for compositional reward transfer. Selected combinations are sources with full expert trajectories; the rest are targets with only expert anchors. Detailed environment settings are provided in Appendix~\ref{appendix:experiments}.
    
    \paragraph{Benchmarks.}
    We evaluate ConTraIRL on four MuJoCo continuous control environments: \texttt{Ant}, \texttt{HalfCheetah}, \texttt{Walker2d}, and \texttt{Swimmer}. For each environment, we independently vary the dynamics and goal factors to construct multiple context combinations. Selected combinations serve as source contexts with full expert trajectories, while the remaining combinations serve as target contexts with only expert anchors. The detailed environment setting can be found in Appendix~\ref{appendix:experiments}.

    \paragraph{Baselines.}
    We select three baselines that capture the main alternatives to ConTraIRL: abstraction based reward transfer, direct matching of expert states, and contrastive reward learning. 
    \textbf{TraIRL} \cite{trairl} transfers rewards across dynamics using a shared abstraction under a fixed goal. It provides the closest comparison for evaluating the benefit of factorizing dynamics and goal information. 
    \textbf{SFM} \cite{sfm} directly matches expert and learner successor features without learning an explicit reward, providing a recent baseline for imitation from expert states. 
    \textbf{PCIL} \cite{PCIL} learns a contrastive representation of expert and learner behavior and defines the reward through abstraction similarity. It provides the closest comparison to our contrastive objective and similarity based reward. A comprehensive baselines comparison can be found in Appendix~\ref{appendix:experiments}.

    \subsection{Benchmark Results}
        \begin{table}[!ht]
            \centering
            \small
            \setlength{\tabcolsep}{0.8mm}

                \begin{tabular}{ccccc}
                    \toprule
                               & Ant               & HalfCheetah       & Walker            & Swimmer           \\
                    \midrule
                    TraIRL     & $ 0.57 \pm 0.21 $ & $ 0.55 \pm 0.25 $ & $ 0.51 \pm 0.22 $ & $ 0.50 \pm 0.28 $ \\
                    SFM        & $ 0.88 \pm 0.09 $ & $ 0.90 \pm 0.05 $ & $ 0.89 \pm 0.06 $ & $ 0.90 \pm 0.04 $ \\
                    PCIL       & $ 0.89 \pm 0.03 $ & $ 0.92 \pm 0.02 $ & $ 0.91 \pm 0.04 $ & $ 0.88 \pm 0.07 $ \\
                    ConTraIRL & $ \textbf{0.93} \pm \textbf{0.02} $ & $ \textbf{0.95} \pm \textbf{0.01} $ & $ \textbf{0.96} \pm \textbf{0.01} $ & $ \textbf{0.97} \pm \textbf{0.02} $ \\
                    \bottomrule
                \end{tabular}
            \caption{Benchmark results on target contexts. For each env, we evaluate on multiple target contexts. Scores are normalized per environment by expert return and then averaged over all target contexts and random seeds (mean $\pm$ std).}
            \label{tab:benchmark_target_avg}
        \end{table}

        \begin{table}[!ht]
            \centering
            \small
            \setlength{\tabcolsep}{0.8mm}
                \begin{tabular}{ccccc}
                    \toprule
                               & Ant               & HalfCheetah       & Walker            & Swimmer           \\
                    \midrule
                    TraIRL     & $ 0.96 \pm 0.01 $ & $ 0.97 \pm 0.00 $ & $ 0.96 \pm 0.02 $ & $ 0.98 \pm 0.01 $ \\
                    ConTraIRL & $ 0.97 \pm 0.01 $ & $ 0.97 \pm 0.01 $ & $ 0.97 \pm 0.01 $ & $ 0.97 \pm 0.02 $ \\
                    \bottomrule
                \end{tabular}
            \caption{Performance when the goal factor is fixed and only dynamics vary. In this setting, TraIRL and ConTraIRL achieve comparable performance, indicating that the primary advantage of ConTraIRL arises in compositional settings.}
            \label{tab:trairl_comparison}
        \end{table}

        Table~\ref{tab:benchmark_target_avg} reports normalized returns in target contexts formed by unseen dynamics and goal pairings. ConTraIRL achieves the highest normalized return across all four environments, demonstrating improved reward recovery for unseen dynamics and goal pairings over the evaluated baselines.

        Table~\ref{tab:trairl_comparison} isolates the role of factorized abstractions by fixing the goal factor. In this setting, reward transfer depends only on dynamics variation, and TraIRL and ConTraIRL achieve comparable performance. The improvement in Table~\ref{tab:benchmark_target_avg} therefore reflects the ability of ConTraIRL to transfer dynamics and goal information separately through its factorized abstractions.


    \subsection{Additional Empirical Results}
        
        \paragraph{Effect of factorization.}
        We compare ConTraIRL with a variant that uses a single encoder for dynamics and goal information, denoted \emph{w/o F}. All other components remain unchanged. Table~\ref{tab:ablation} shows that removing factorization substantially reduces target performance. A single abstraction must represent both factors and generalize directly to an unseen combination. Factorized abstractions instead allow the information associated with each target factor to transfer separately from the source contexts.

        \paragraph{Effect of phase consistency.}
        We remove the phase condition when constructing pairs for \(\mathcal L_E\) and \(\mathcal L_\pi\), denoted \emph{w/o PC}. Without phase consistency, states sharing the same factor may receive similar abstractions even when they belong to different phases. The resulting decrease in Table~\ref{tab:ablation} shows that preserving phase dependent variation is important for reward learning.

        \begin{table}[!ht]
            \centering
            \small
            \setlength{\tabcolsep}{1mm}
            
            \begin{tabular}{lccc}
                \toprule
                             & Ant               & HalfCheetah       & Walker            \\
                \midrule
                ConTraIRL w/o F  & $ 0.27 \pm 0.12 $ & $ 0.59 \pm 0.16 $ & $ 0.58 \pm 0.13 $ \\
                ConTraIRL w/o PC & $ 0.82 \pm 0.06 $ & $ 0.88 \pm 0.07 $ & $ 0.90 \pm 0.04 $ \\
                ConTraIRL          & $ \textbf{0.93} \pm \textbf{0.02} $ & $ \textbf{0.95} \pm \textbf{0.01} $ & $ \textbf{0.96} \pm \textbf{0.01} $ \\
                \bottomrule
            \end{tabular}
            \caption{Ablation results on target contexts. ``w/o F'' removes factorized encoders; ``w/o PC'' removes phase consistency.}
            \label{tab:ablation}
        \end{table}

        \paragraph{Effect of the ratio of target anchors.}
        We vary the ratio of target anchors while keeping all other conditions fixed. Table~\ref{tab:amount expert} shows that performance decreases for all methods as fewer anchors are available, although ConTraIRL remains strongest whenever anchors are provided. Without anchors, all methods perform poorly; fully zero-shot transfer is outside the scope of this work.
        
        \begin{table}[!ht]
            \centering
            \small
            \setlength{\tabcolsep}{0.8mm}            
            \begin{tabular}{ccccc}
                \toprule
                Ratio &       20\%        &              10\% &               5\%  & 0\% \\
                \midrule
                TraIRL        & $ 0.55 \pm 0.25 $ & $ 0.05 \pm 0.19 $ & $ 0.04 \pm 0.03 $ & $ 0.05 \pm 0.08$\\
                SFM           & $ 0.90 \pm 0.05 $ & $ 0.31 \pm 0.10 $ & $ 0.05 \pm 0.04 $ & $0.06 \pm 0.11$\\
                PCIL          & $ 0.89 \pm 0.03 $ & $ 0.37 \pm 0.11 $ & $ 0.19 \pm 0.06 $ & $0.03 \pm 0.12$\\
                ConTraIRL  & $ \textbf{0.95} \pm \textbf{0.01} $ & $ \textbf{0.53} \pm \textbf{0.05} $ & $ \textbf{0.27} \pm \textbf{0.08} $  & $ 0.04 \pm 0.09$\\
                \bottomrule
            \end{tabular}
            \caption{Effect of the ratio of target anchors in \texttt{HalfCheetah}. Normalized return in target contexts.}
            \label{tab:amount expert}
        \end{table}

\begin{figure}[!ht]
            \centering
            \includegraphics[width=1.0\columnwidth]{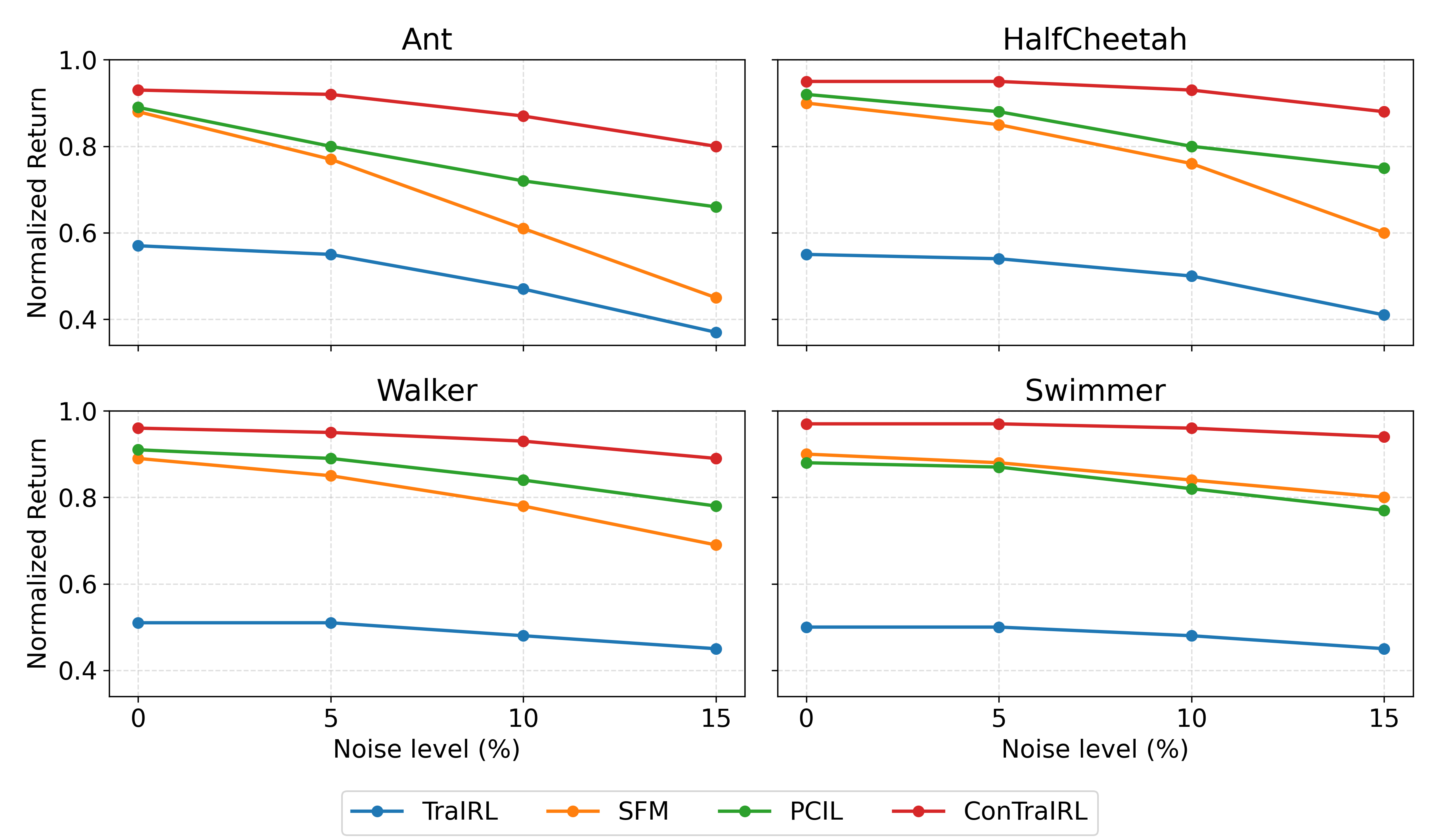}
            \caption{Effect of noise in factor labels. ConTraIRL degrades more slowly than baselines across all environments, indicating stronger robustness to noisy factors.}
            \label{fig:noise_performance}
        \end{figure}

        \paragraph{Effect of noise in factor labels.}
        Although our formulation assumes that factor labels are provided by the environment, we evaluate sensitivity to randomly noisy labels. Figure~\ref{fig:noise_performance} shows that ConTraIRL degrades more gradually than the baselines as the corruption rate increases. This result indicates that the contrastive objectives tolerate moderate errors in factor conditioning and pair construction. In practice, dynamics labels may be unavailable and must be inferred from trajectories. We evaluate this setting in Appendix~\ref{appendix:experiments}.
        
    \subsection{Analysis on the Learned Abstractions}

        \begin{figure}[!ht]
            \centering
            \includegraphics[width=1.0\columnwidth]{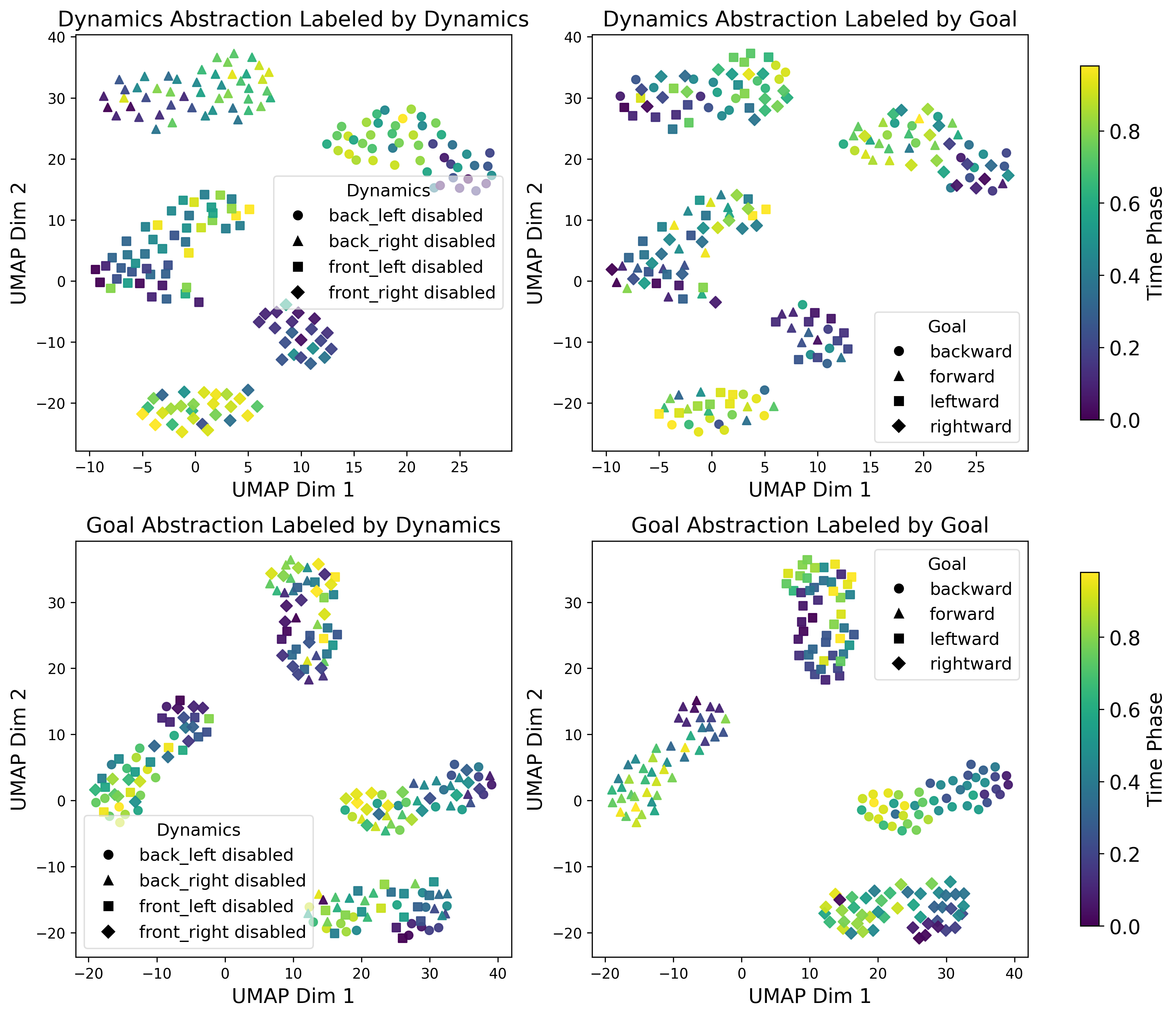}
            \caption{UMAP visualization of the learned abstractions. Dynamics abstractions separate dynamics while mixing goals, and goal abstractions show the reverse. For visualization, ordered discrete phases are mapped to \([0,1]\).}
            \label{fig:abstractions_visualization}
        \end{figure}
        
        We use UMAP \cite{umap} to visualize the learned abstractions to examine whether the factorized encoders recover cross factor invariant, phase-consistent, and structurally coherent abstractions.

        \paragraph{Cross factor invariance.}
        In the dynamics abstraction, states separate by dynamics while goal labels remain intermixed (top row); the goal abstraction exhibits the reverse pattern (bottom row). To quantify this property, we train lightweight two-layer MLP probes on the frozen abstractions to predict dynamics and goal labels. The dynamics abstraction \(z_d\) achieves \(94.8\%\) accuracy for dynamics prediction and \(43.6\%\) for goal prediction, while the goal abstraction \(z_g\) achieves \(97.1\%\) accuracy for goal prediction and \(34.5\%\) for dynamics prediction. The chance accuracies for dynamics and goal prediction are \(20.0\%\) and \(25.0\%\), respectively. The large accuracy gap shows that each abstraction is substantially more predictive of its relevant factor. The remaining predictability of the other factor primarily reflects the nonzero margin for negative pairs in Eq.\eqref{eq:expert_pair_loss}, which permits residual similarity instead of enforcing strict separation. These results are therefore consistent with the intended conditional cross-factor invariance. 

        \paragraph{Phase consistency.}
        Within each factor group, colors indicate the ordered discrete phase labels mapped to \([0,1]\) for visualization. States are organized according to their phases, indicating that the abstractions retain phase information even though phase labels are not provided to the encoders.

        \paragraph{Expert and learner abstraction alignment.} Table~\ref{tab:abstraction_similarity} reports mean cosine similarities for expert pairs and expert and learner pairs computed using the final encoders. The first row shows high similarity between expert states sharing the relevant factor and phase but differing in the other factor, indicating strong cross factor invariance. The second and third rows show that, under the fixed final encoders, same phase similarity increases from the early to the final policy, indicating that policy optimization aligns learner states with the corresponding expert abstractions in both factor spaces. The fourth row shows lower similarity to expert states from different phases, indicating that the alignment remains phase selective rather than resulting from representation collapse.

        \begin{table}[!ht]
        \centering
        \small
        \setlength{\tabcolsep}{1mm}
        \begin{tabular}{lcc}
        \toprule
        Comparison & Dynamics & Goal \\
        \midrule
        Expert pair, same phase          & 0.91 $\pm$ 0.05 & 0.94 $\pm$ 0.02 \\
        Early learner, same phase        & 0.17 $\pm$ 0.12 & 0.11 $\pm$ 0.08 \\
        Final learner, same phase        & 0.85 $\pm$ 0.03 & 0.89 $\pm$ 0.04 \\
        Final learner, different phase   & 0.54 $\pm$ 0.21 & 0.59 $\pm$ 0.10 \\
        \bottomrule
        \end{tabular}
        \caption{Mean cosine similarities to source expert abstractions under the final encoders. Rows 1 to 3 use same-factor, same-phase references; in Row 1, the expert state and reference come from contexts differing in the other factor. Row 4 uses a same-factor, different-phase reference.}
        \label{tab:abstraction_similarity}
        \end{table}

        Together, these results show that ConTraIRL learns factorized abstractions that exhibit cross-factor invariance and phase consistency, while learner abstractions become increasingly aligned with expert behavior as the policy improves. These support reward transfer to target contexts.

    \subsection{Evaluation on Simulated Robotics Tasks}

        We further evaluate ConTraIRL in Assistive Gym, which provides a more challenging simulated robotic setting with high dimensional control, human robot interaction, and constrained motion around physical obstacles~\cite{assistive-gym}. We consider assistive goals including feeding, drinking, and scratching, and construct different dynamics contexts by varying the obstacle configuration around the robot and human. These settings require ConTraIRL to transfer goal specific behavior across substantially different trajectories and state occupancies.
        
        \begin{figure}[!ht]
            \centering
            \begin{subfigure}[t]{0.32\columnwidth}
                \centering
                \includegraphics[width=\linewidth]{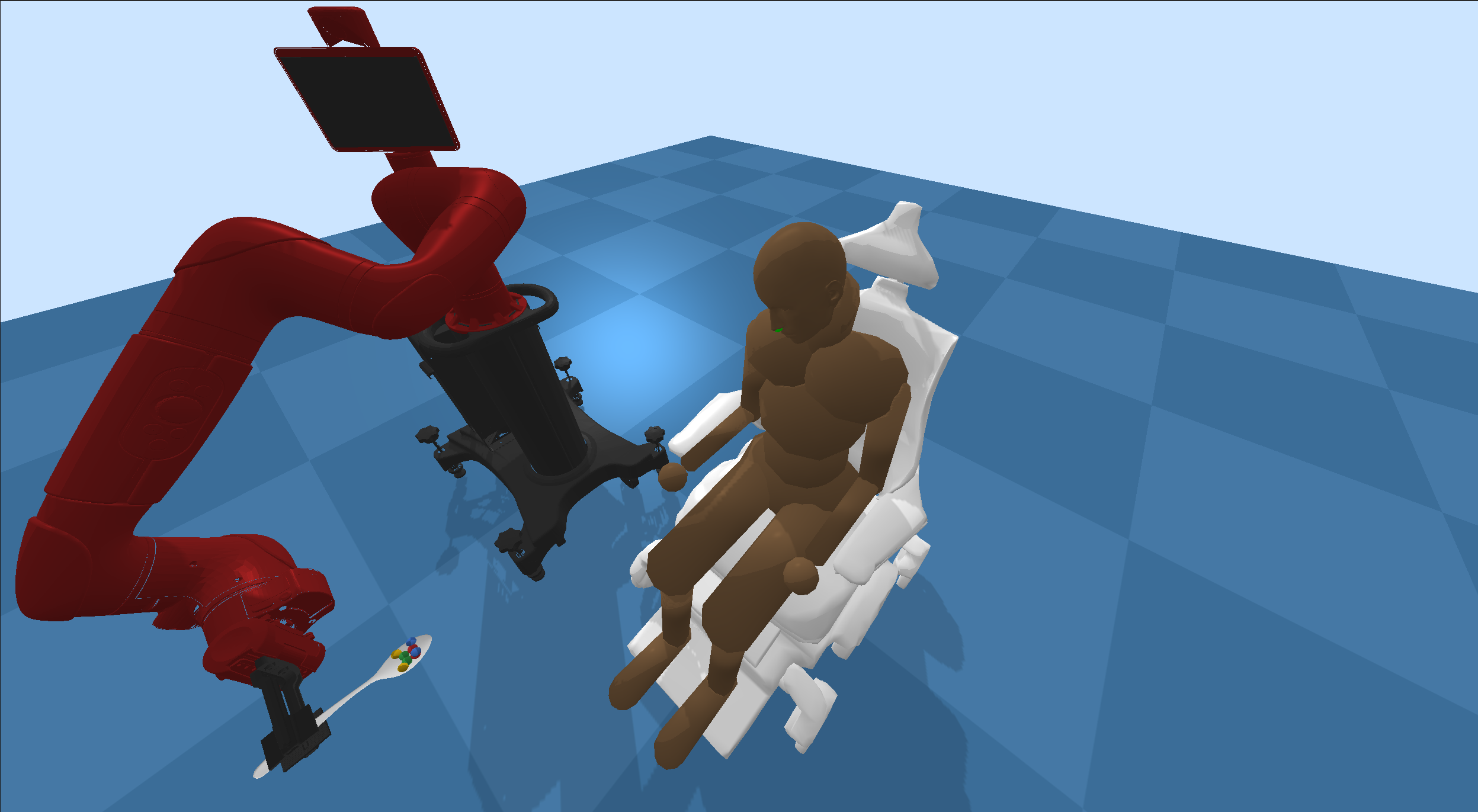}
                \caption{}
            \end{subfigure}
            \hfill
            \begin{subfigure}[t]{0.32\columnwidth}
                \centering
                \includegraphics[width=\linewidth]{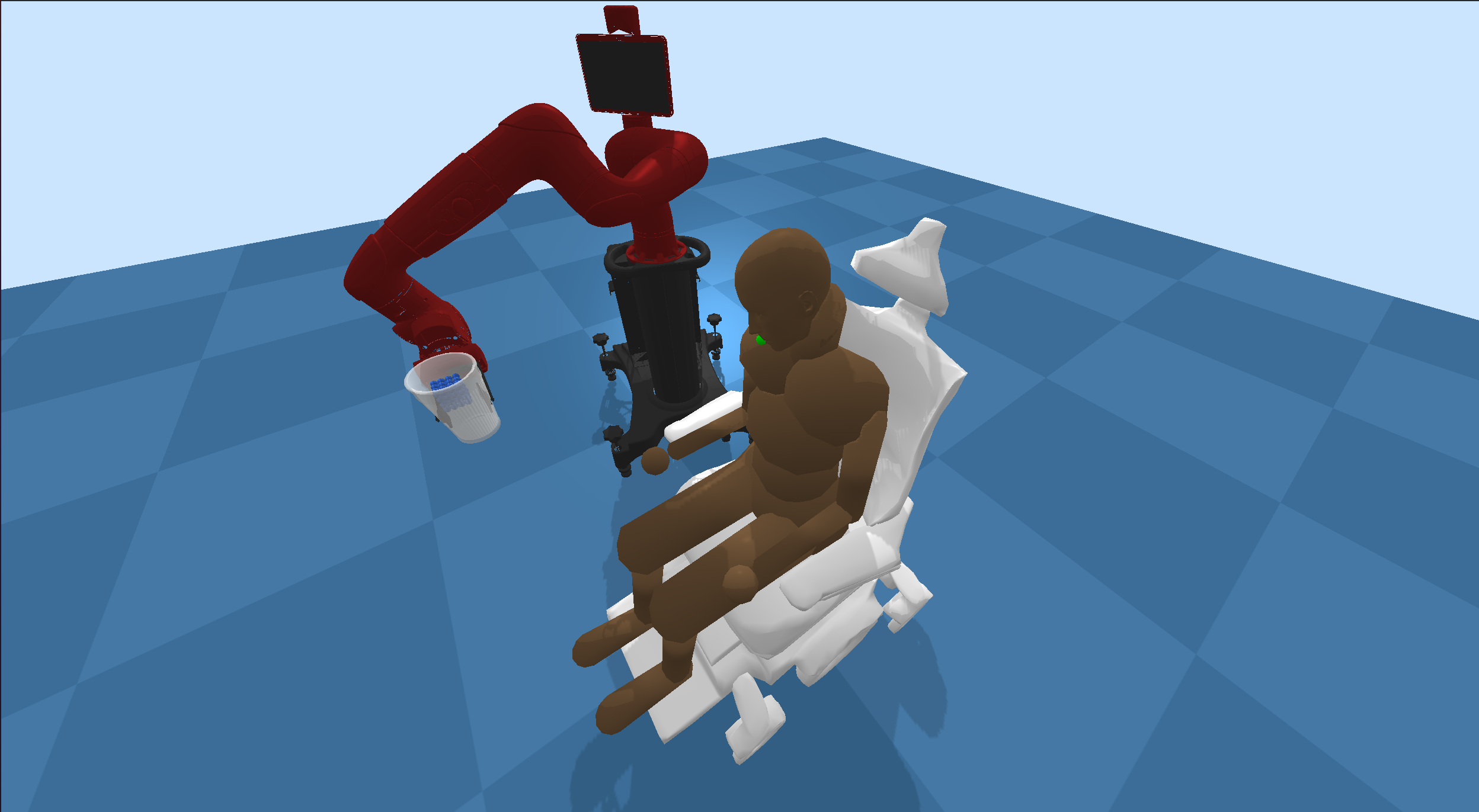}
                \caption{}
            \end{subfigure}
            \hfill
            \begin{subfigure}[t]{0.32\columnwidth}
                \centering
                \includegraphics[width=\linewidth]{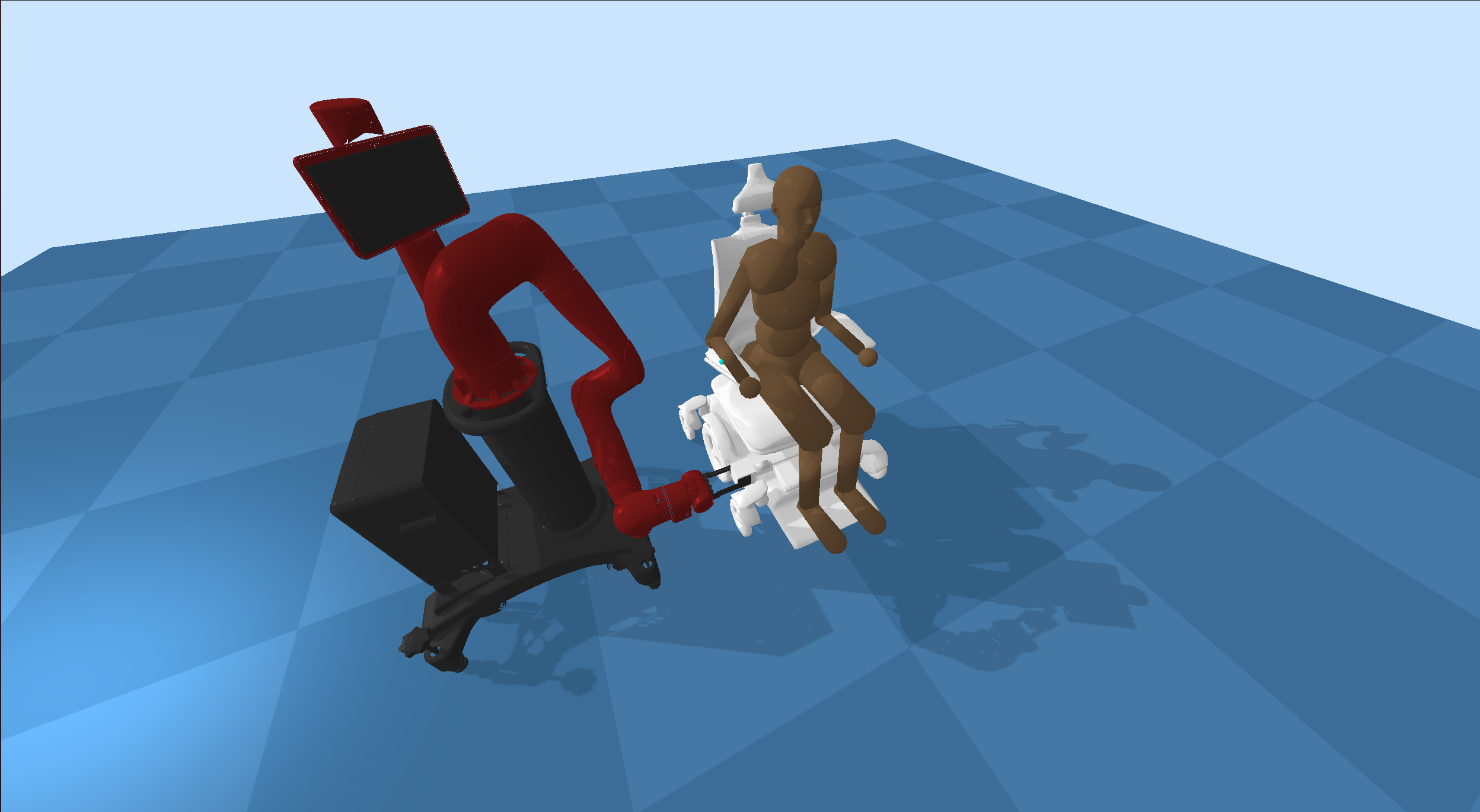}
                \caption{}
            \end{subfigure}
            \caption{Assistive Gym environments. (a) Feeding task. (b) Drinking task. (c) Scratch Itch task.}
            \label{fig:assistive_gym}
        \end{figure}

        \begin{table}[!ht]
            \centering
            \small
            \begin{tabular}{lcc}
                \toprule
                Method & Normalized Return & Success Rate \\
                \midrule
                TraIRL    & 0.27 $\pm$ 0.11 & 0.04 $\pm$ 0.02 \\
                SFM       & 0.61 $\pm$ 0.18 & 0.31 $\pm$ 0.08 \\
                PCIL      & 0.43 $\pm$ 0.09 & 0.25 $\pm$ 0.06 \\
                ConTraIRL & 0.70 $\pm$ 0.08 & 0.62 $\pm$ 0.09 \\
                \bottomrule
            \end{tabular}
            \caption{Performance on target contexts with unseen combinations of obstacle configurations and assistive goals.}
            \label{tab:assistive_results}
        \end{table}

        Table~\ref{tab:assistive_results} shows that ConTraIRL achieves the highest normalized return and success rate on the unseen obstacle--goal combinations. Compared with the baselines, ConTraIRL more effectively transfers goal-specific behavior while adapting to the distinct trajectories induced by each obstacle configuration. These results indicate that the factorized abstractions remain effective in a more complex robotic setting where dynamics variations produce substantial changes in feasible behavior and state occupancy.

\section{Conclusion}

    We introduced ConTraIRL for compositional reward transfer to target contexts that combine previously observed dynamics and goal factors in a new pairing. Its key novelty is factorized occupancy matching: for each phase, ConTraIRL matches expert and learner occupancies separately for the dynamics and goal factors, rather than matching a single occupancy for their joint context. Cross factor invariance allows each abstraction to transfer across changes in the other factor, while phase consistency preserves behavioral progress. Our analysis shows that minimizing the ConTraIRL objective is equivalent to match the expert dynamics and goal occupancies, enabling transfer to their unseen combination. Target anchors provide additional calibration when the source and target occupancies differ.
    Experiments across four MuJoCo environments show that ConTraIRL consistently improves performance on unseen combinations of dynamics and goals. Ablations and probing experiments verify the contributions of factorization, phase consistency, and target anchors.



\newpage
\setcounter{secnumdepth}{2}

\appendix

\onecolumn
\section{Algorithm} \label{appendix:algorithm}

\begin{algorithm}[!ht]
\caption{ConTraIRL Training Procedure}
\label{alg:contrairl}
\textbf{Input}: Source expert trajectories $\mathcal{D}_E$, source contexts $\mathcal{C}_{\mathrm{src}}$, target anchors $\mathcal{B}_{\mathrm{tar}}^E$, target context $c_{\mathrm{tar}}$, and phase space $\Psi$\\
\textbf{Parameter}: Training epochs $N$, encoder steps $K_{\mathrm{enc}}$, and policy steps $K_{\mathrm{policy}}$\\
\textbf{Output}: Learned policies $\{\widehat{\pi}_c\}$ and reward $\widehat{R}$
\begin{algorithmic}[1]
\STATE Set $\mathcal{C}_{\mathrm{train}}=\mathcal{C}_{\mathrm{src}}\cup\{c_{\mathrm{tar}}\}$
\STATE Initialize shared encoders $\Phi_d,\Phi_g$ and policy $\pi_{\theta_c}$ for each $c\in\mathcal{C}_{\mathrm{train}}$
\FOR{$n=1,\ldots,N$}
    \FOR{each context $c\in\mathcal{C}_{\mathrm{train}}$}
        \STATE Collect phase labeled learner rollouts $\mathcal{D}_\pi^c$ using $\pi_{\theta_c}$
    \ENDFOR
    \STATE Set $\mathcal{D}_\pi=\bigcup_{c\in\mathcal{C}_{\mathrm{train}}}\mathcal{D}_\pi^c$
    \STATE \textbf{------------ Contrastive encoder update ------------}
    \FOR{$k=1,\ldots,K_{\mathrm{enc}}$}
        \STATE Sample source expert pairs and construct $y_{ij}^d$ and $y_{ij}^g$
        \STATE Sample target anchor and source expert pairs matched by factor and phase
        \STATE Compute the expert abstraction loss $\mathcal{L}_E$ using Eq.~(8)
        \STATE Sample expert positives and learner negatives matched by dynamics and phase for $\Phi_d$
        \STATE Sample expert positives and learner negatives matched by goal and phase for $\Phi_g$
        \STATE Compute the expert and learner separation loss $\mathcal{L}_\pi$ using Eq.~(9)
        \STATE Update $\Phi_d$ and $\Phi_g$ by minimizing $\mathcal{L}_E+\mathcal{L}_\pi$
    \ENDFOR
    \STATE \textbf{Expert abstraction update}
    \FOR{each observed dynamics factor $d$ and phase $\psi$}
        \STATE Compute $z_d^E(\psi)$ from $\rho_E(\cdot\mid d,\psi)$ using Eq.~(2)
    \ENDFOR
    \FOR{each observed goal factor $g$ and phase $\psi$}
        \STATE Compute $z_g^E(\psi)$ from $\rho_E(\cdot\mid g,\psi)$ using Eq.~(2)
    \ENDFOR
    \STATE \textbf{------------ Policy optimization ------------}
    \FOR{each context $c=(d,g)\in\mathcal{C}_{\mathrm{train}}$}
        \FOR{$k=1,\ldots,K_{\mathrm{policy}}$}
            \STATE Collect $\tau_\pi^c$ using $\pi_{\theta_c}$
            \STATE Compute $\widehat{R}(s,d,g,\psi)$ using Eq.~(3)
            \STATE Update $\pi_{\theta_c}$ using Soft Actor Critic with $\widehat{R}$
        \ENDFOR
    \ENDFOR
\ENDFOR
\STATE \textbf{return} $\{\widehat{\pi}_c=\pi_{\theta_c}\}_{c\in\mathcal{C}_{\mathrm{train}}}$ and $\widehat{R}$
\end{algorithmic}
\end{algorithm}

\section{Proof} \label{appendix:proof}
    \newtheorem*{repeatedlemma}{Lemma~\ref{lemma:objective_equivalence}}
    \begin{repeatedlemma}
        Under a first-order Taylor approximation of the log-sum-exp terms in \(\mathcal L_\pi\), minimizing the resulting objective with respect to \(\Phi_d\) and \(\Phi_g\) is equivalent to the inner reward maximization in Eq.~\eqref{eq:contrairl_al_objective}, namely, maximizing the factorized expert-learner reward gap \(\Delta_d(\Phi_d,\pi)+\Delta_g(\Phi_g,\pi)\).
    \end{repeatedlemma}
    Following PCIL~\cite{PCIL}, we analyze the case of a single learner negative; multiple negatives only change the positive scaling factor under the same approximation.

    \begin{proof}
    We first consider the dynamics component of \(\mathcal L_\pi\). Let
    \[
    a_d=(z_{d,i}^E)^\top z_{d,j}^E,
    \qquad
    b_d=(z_{d,i}^E)^\top z_d^\pi,
    \]
    where \(z_{d,i}^E\) and \(z_{d,j}^E\) are independently sampled expert abstractions and \(z_d^\pi\) is a learner abstraction with the same dynamics and phase. With one learner negative, the dynamics loss can be written as
    \[
    \mathcal L_\pi^d
    =
    \mathbb E\left[
    \log\left(
    1+\exp\left(\frac{b_d-a_d}{\tau}\right)
    \right)
    \right].
    \]
    Using the first-order Taylor approximation
    \[
    \log(1+\exp(x))
    \approx
    \log 2+\frac{x}{2}
    \]
    around \(x=0\), and dropping the additive constant and positive scaling factor, we obtain
    \[
    \mathcal L_\pi^d
    \propto
    \mathbb E[b_d-a_d].
    \]
    By independence of the two expert samples,
    \[
    \begin{aligned}
    \mathbb E[a_d]
    &=
    \left(
    \mathbb E_{s\sim\rho_E(\cdot\mid d,\psi)}
    [\Phi_d(s,d)]
    \right)^\top
    \left(
    \mathbb E_{s\sim\rho_E(\cdot\mid d,\psi)}
    [\Phi_d(s,d)]
    \right) \\
    &=
    \mathbb E_{s\sim\rho_E(\cdot\mid d,\psi)}
    \left[
    \Phi_d(s,d)^\top z_d^E(\psi)
    \right].
    \end{aligned}
    \]
    Similarly, independence of the expert and learner samples gives
    \[
    \mathbb E[b_d]
    =
    \mathbb E_{s\sim\rho_\pi(\cdot\mid d,\psi)}
    \left[
    \Phi_d(s,d)^\top z_d^E(\psi)
    \right].
    \]
    Therefore,
    \[
    \mathcal L_\pi^d
    \propto
    -\Delta_d(\Phi_d,\pi).
    \]
    Applying the same argument to the goal component gives
    \[
    \mathcal L_\pi^g
    \propto
    -\Delta_g(\Phi_g,\pi).
    \]
    Consequently,
    \[
    \mathcal L_\pi
    \propto
    -\left[
    \Delta_d(\Phi_d,\pi)+\Delta_g(\Phi_g,\pi)
    \right].
    \]
    Thus, minimizing the approximated \(\mathcal L_\pi\) with respect to \(\Phi_d\) and \(\Phi_g\) is equivalent to maximizing \(\Delta_d(\Phi_d,\pi)+\Delta_g(\Phi_g,\pi)\), which is the inner reward maximization in Eq.~\eqref{eq:contrairl_al_objective}.
    \end{proof}

    \newtheorem*{repeatedtheorem}{Theorem~\ref{theorem:occupancy_matching}}
    \begin{repeatedtheorem}
        The factor-conditioned contrastive discrepancy satisfies
        \begin{equation}
        \frac{1}{4}
        \left(D_{\mathrm{TV}}^d+D_{\mathrm{TV}}^g\right)
        \leq
        \mathcal D_{\mathrm{ConTra}}(\rho_E,\rho_\pi)
        \leq
        2
        \left(D_{\mathrm{TV}}^d+D_{\mathrm{TV}}^g\right).
        \nonumber
        \end{equation}
    \end{repeatedtheorem}

    \begin{proof}
    Following the scalar reduction used in PCIL~\cite{PCIL}, we first establish the result for a single factor. For \(f\in\{d,g\}\), let
    \[
    (P_f,Q_f)
    =
    \begin{cases}
    \bigl(\rho_E(\cdot\mid d,\psi),\rho_\pi(\cdot\mid d,\psi)\bigr), & f=d,\\
    \bigl(\rho_E(\cdot\mid g,\psi),\rho_\pi(\cdot\mid g,\psi)\bigr), & f=g,
    \end{cases}
    \]
    and suppress the fixed factor input of \(\Phi_f\). Define
    \[
    v_f=\mathbb E_{s\sim P_f}[\Phi_f(s)].
    \]
    The corresponding factor-conditioned reward gap is
    \[
    \Delta_f(\Phi_f,\pi)
    =
    v_f^\top
    \left(
    \mathbb E_{s\sim P_f}[\Phi_f(s)]
    -
    \mathbb E_{s\sim Q_f}[\Phi_f(s)]
    \right).
    \]
    
    \paragraph{Scalar reduction.}
    Choose an orthonormal matrix \(M_f\) such that
    \[
    M_fv_f=\|v_f\|_2e_1,
    \]
    where \(e_1\) is the first standard basis vector. Define
    \[
    \alpha_f=\|v_f\|_2,
    \qquad
    g_f(s)=e_1^\top M_f\Phi_f(s).
    \]
    Since \(M_f\) preserves norms and \(\|\Phi_f(s)\|_2\leq1\), we have \(|g_f(s)|\leq1\). Moreover,
    \[
    \mathbb E_{s\sim P_f}[g_f(s)]
    =
    e_1^\top M_f\mathbb E_{s\sim P_f}[\Phi_f(s)]
    =
    e_1^\top M_fv_f
    =
    \alpha_f.
    \]
    Using invariance of dot products under orthonormal transformations,
    \[
    \begin{aligned}
    \Delta_f(\Phi_f,\pi)
    &=
    (M_fv_f)^\top
    M_f
    \left(
    \mathbb E_{P_f}[\Phi_f]
    -
    \mathbb E_{Q_f}[\Phi_f]
    \right)\\
    &=
    \alpha_f
    \left(
    \mathbb E_{P_f}[g_f]
    -
    \mathbb E_{Q_f}[g_f]
    \right)\\
    &=
    \alpha_f
    \int_{\mathcal S}
    g_f(s)\bigl(P_f(s)-Q_f(s)\bigr)\,ds.
    \end{aligned}
    \]
    
    \paragraph{Lower bound.}
    Let
    \[
    S_f=\{s\in\mathcal S:P_f(s)\geq Q_f(s)\},
    \]
    and denote
    \[
    \delta_f
    =
    D_{\mathrm{TV}}(P_f,Q_f)
    =
    P_f(S_f)-Q_f(S_f).
    \]
    Consider the scalar function
    \[
    g_f^\star(s)
    =
    \begin{cases}
    1, & s\in S_f,\\
    \frac12, & s\notin S_f,
    \end{cases}
    \]
    and the encoder
    \[
    \Phi_f^\star(s)=g_f^\star(s)e_1.
    \]
    Since \(\|\Phi_f^\star(s)\|_2\leq1\), this encoder belongs to the admissible encoder class. Its expert mean satisfies
    \[
    \begin{aligned}
    \alpha_f^\star
    &=
    \mathbb E_{P_f}[g_f^\star]\\
    &=
    P_f(S_f)+\frac12P_f(S_f^c)\\
    &=
    \frac{1+P_f(S_f)}{2}\\
    &\geq
    \frac12.
    \end{aligned}
    \]
    Furthermore,
    \[
    \begin{aligned}
    \mathbb E_{P_f}[g_f^\star]
    -
    \mathbb E_{Q_f}[g_f^\star]
    &=
    P_f(S_f)-Q_f(S_f)
    +
    \frac12\left(P_f(S_f^c)-Q_f(S_f^c)\right)\\
    &=
    \frac12\left(P_f(S_f)-Q_f(S_f)\right)\\
    &=
    \frac12\delta_f.
    \end{aligned}
    \]
    Therefore,
    \[
    \Delta_f(\Phi_f^\star,\pi)
    =
    \alpha_f^\star
    \left(
    \mathbb E_{P_f}[g_f^\star]
    -
    \mathbb E_{Q_f}[g_f^\star]
    \right)
    \geq
    \frac14\delta_f.
    \]
    Since the maximum over \(\Phi_f\) is at least the value achieved by \(\Phi_f^\star\),
    \[
    \max_{\Phi_f}\Delta_f(\Phi_f,\pi)
    \geq
    \frac14D_{\mathrm{TV}}(P_f,Q_f).
    \]
    
    \paragraph{Upper bound.}
    For any admissible encoder,
    \[
    \alpha_f
    =
    \left\|
    \mathbb E_{P_f}[\Phi_f(s)]
    \right\|_2
    \leq
    \mathbb E_{P_f}[\|\Phi_f(s)\|_2]
    \leq
    1.
    \]
    Together with \(|g_f(s)|\leq1\), this gives
    \[
    \begin{aligned}
    \Delta_f(\Phi_f,\pi)
    &\leq
    \left|
    \Delta_f(\Phi_f,\pi)
    \right|\\
    &=
    \alpha_f
    \left|
    \int_{\mathcal S}
    g_f(s)\bigl(P_f(s)-Q_f(s)\bigr)\,ds
    \right|\\
    &\leq
    \int_{\mathcal S}
    |g_f(s)|\,|P_f(s)-Q_f(s)|\,ds\\
    &\leq
    \int_{\mathcal S}
    |P_f(s)-Q_f(s)|\,ds\\
    &=
    2D_{\mathrm{TV}}(P_f,Q_f).
    \end{aligned}
    \]
    Maximizing over \(\Phi_f\) yields
    \[
    \max_{\Phi_f}\Delta_f(\Phi_f,\pi)
    \leq
    2D_{\mathrm{TV}}(P_f,Q_f).
    \]
    Combining the factor-wise lower and upper bounds,
    \[
    \frac14D_{\mathrm{TV}}(P_f,Q_f)
    \leq
    \max_{\Phi_f}\Delta_f(\Phi_f,\pi)
    \leq
    2D_{\mathrm{TV}}(P_f,Q_f).
    \]
    
    \paragraph{Combining the two factors.}
    Because \(\Phi_d\) and \(\Phi_g\) are optimized independently,
    \[
    \begin{aligned}
    D_{\mathrm{ConTra}}(\rho_E,\rho_\pi)
    &=
    \max_{\Phi_d,\Phi_g}
    \left[
    \Delta_d(\Phi_d,\pi)+\Delta_g(\Phi_g,\pi)
    \right]\\
    &=
    \max_{\Phi_d}\Delta_d(\Phi_d,\pi)
    +
    \max_{\Phi_g}\Delta_g(\Phi_g,\pi).
    \end{aligned}
    \]
    Applying the factor-wise bounds to the dynamics and goal occupancies and summing gives
    \[
    \frac14
    \left(
    D_{\mathrm{TV}}^d+D_{\mathrm{TV}}^g
    \right)
    \leq
    D_{\mathrm{ConTra}}(\rho_E,\rho_\pi)
    \leq
    2
    \left(
    D_{\mathrm{TV}}^d+D_{\mathrm{TV}}^g
    \right).
    \]
    \end{proof}

\section{Experiments} \label{appendix:experiments}
    \subsection{Benchmark Environment Setting}
    Each target context is excluded from the source demonstration set, although limited anchor states from that context are available during training. Its dynamics and goal factors each appear in at least one source context, so transfer requires recombining source-supported factors rather than extrapolating to unseen factors. All methods use identical context partitions and target-anchor budgets. The resulting partitions contain 44 source and 12 target contexts for Ant, 9 source and 3 target contexts each for HalfCheetah and Walker2d, and 3 source and 3 target contexts for Swimmer. The partitions are listed in Tables~\ref{tab:ant_contexts}, \ref{tab:halfcheetah_walker_contexts}, and~\ref{tab:swimmer_contexts}.
    The environment reward is used only to train expert policies and report evaluation returns; it is not provided to the IRL methods or their learned policies.

    \paragraph{Dynamics and goal construction.} We construct the dynamics factors by varying which actuators remain functional. For Ant, only the indicated one, two, or three legs remain functional. For HalfCheetah and Walker2d, only the indicated leg remains functional, except in the full-functionality setting. For Swimmer, only joint 1, joint 2, or all joints remain functional. HalfCheetah and Walker2d additionally distinguish walking and running as lower- and higher-speed behaviors under each dynamics configuration. The goal factors specify the desired direction of motion. Ant uses forward, backward, leftward, and rightward goals, while Swimmer uses forward and backward goals.  The task rewards are used only to train expert policies and report evaluation returns; they are not provided to the IRL methods.

    \paragraph{Expert data and target anchors.} For each source context, every method receives one complete state-only expert trajectory with a phase label for each state. Unless otherwise specified, the target anchors are obtained by randomly sampling \(20\%\) of the states from one phase-labeled target expert trajectory at the beginning of training. These anchors remain fixed throughout training, and no other target expert states are accessible. Expert actions and task rewards are not provided. The dynamics, goal, and phase labels are available to all methods, which collect learner rollouts while optimizing their learned rewards. All methods use identical source demonstrations and target anchors.
    
    \begin{table}[!ht]
        \centering
        \small
        \begin{tabular}{ccc}
            \toprule
            Dynamics factor              & Goal factor& Split  \\
            \midrule
            joint 1  & forward   & Target \\
            joint 1  & backward  & Source \\
            \midrule
            joint 2  & forward   & Source \\
            joint 2 &  backward  & Target \\
            \midrule
            all joints   & forward      & Target \\
            all joints   & backward     & Source \\
            \bottomrule
        \end{tabular}
        \caption{Context configurations for the Swimmer environment.}
        \label{tab:swimmer_contexts}
    \end{table}

    \begin{table*}[!ht]
        \centering
        \small
        \begin{minipage}{0.48\textwidth}
            \centering
            \textbf{HalfCheetah}

            \begin{tabular}{ccc}
                \toprule
                Dynamics factor & Goal factor & Split \\
                \midrule
                front leg  & run forward   & Source \\
                front leg  & run backward  & Source \\
                front leg  & walk forward  & Target \\
                front leg  & walk backward & Source \\
                \midrule
                back leg   & run forward   & Source \\
                back leg   & run backward  & Target \\
                back leg   & walk forward  & Source \\
                back leg   & walk backward & Source \\
                \midrule
                Full functionality & run forward   & Target \\
                Full functionality & run backward  & Source \\
                Full functionality & walk forward  & Source \\
                Full functionality & walk backward & Source \\
                \bottomrule
            \end{tabular}
        \end{minipage}
        \hfill
        \begin{minipage}{0.48\textwidth}
        \centering
        \textbf{Walker}

        \begin{tabular}{ccc}
        \toprule
        Dynamics factor & Goal factor & Split \\
        \midrule
        left leg  & run forward   & Source \\
        left leg  & run backward  & Source \\
        left leg  & walk forward  & Target \\
        left leg  & walk backward & Source \\
        \midrule
        right leg & run forward   & Source \\
        right leg & run backward  & Target \\
        right leg & walk forward  & Source \\
        right leg & walk backward & Source \\
        \midrule
        Full functionality & run forward   & Target \\
        Full functionality & run backward  & Source \\
        Full functionality & walk forward  & Source \\
        Full functionality & walk backward & Source \\
        \bottomrule
        \end{tabular}
        \end{minipage}
        \caption{Context configurations for the HalfCheetah and Walker environments.}
        \label{tab:halfcheetah_walker_contexts}
    \end{table*}

        \begin{table*}[!ht]
            \centering
            \small
            \begin{minipage}{0.48\textwidth}
                \centering
                \begin{tabular}{ccc}
                    \toprule
                    Dynamics factor & Goal factor & Split \\
                    \midrule
                    back right  & forward   & Target \\
                    back right  & leftward  & Source \\
                    back right  & rightward & Source \\
                    back right  & backward  & Source \\
                    \midrule
                    front left  & forward   & Source \\
                    front left  & backward  & Target \\
                    front left  & rightward & Source \\
                    front left  & leftward  & Source \\ 
                    \midrule
                    front right & forward   & Source \\
                    front right & backward  & Source \\
                    front right & rightward & Target \\
                    front right & leftward  & Source \\
                    \midrule
                    back left   & forward   & Source \\
                    back left   & backward  & Source \\
                    back left   & rightward & Source \\
                    back left   & leftward  & Target \\
                    \midrule
                    back right \& front left & forward   & Target \\
                    back right \& front left & backward  & Source \\
                    back right \& front left & leftward  & Source \\
                    back right \& front left & rightward & Source \\
                    \midrule
                    back right \& front right & forward   & Source \\
                    back right \& front right & backward  & Target \\
                    back right \& front right & leftward  & Source \\
                    back right \& front right & rightward & Source \\
                    \midrule
                    back right \& back left & forward   & Source \\
                    back right \& back left & backward  & Source \\
                    back right \& back left & leftward  & Target \\
                    back right \& back left & rightward & Source \\
                    \bottomrule
                \end{tabular}
            \end{minipage}
            \hfill
            \begin{minipage}{0.48\textwidth}
                \centering
                \begin{tabular}{ccc}
                    \toprule
                    Dynamics factor & Goal factor & Split \\
                    \midrule
                    front left \& front right  & forward   & Source \\
                    front left \& front right  & backward  & Source \\
                    front left \& front right  & leftward  & Source \\
                    front left \& front right  & rightward & Target \\
                    \midrule
                    front left \& back left  & forward   & Source \\
                    front left \& back left  & backward  & Source \\
                    front left \& back left  & leftward  & Source \\
                    front left \& back left  & rightward & Source \\
                    \midrule
                    front right \& back left  & forward   & Source \\
                    front right \& back left  & backward  & Source \\
                    front right \& back left  & leftward  & Source \\
                    front right \& back left  & rightward & Source \\
                    \midrule
                    front left \& front right \& back left  & forward    & Target \\
                    front left \& front right \& back left  & backward   & Source \\
                    front left \& front right \& back left  & leftward   & Source \\
                    front left \& front right \& back left  & rightward  & Source \\
                    \midrule
                    back right \& front right \& back left  & forward    & Source \\
                    back right \& front right \& back left  & backward   & Target \\
                    back right \& front right \& back left  & leftward   & Source \\
                    back right \& front right \& back left  & rightward  & Source \\
                    \midrule
                    back right \& front left \& back left  & forward    & Source \\
                    back right \& front left \& back left  & backward   & Source \\
                    back right \& front left \& back left  & leftward   & Target \\
                    back right \& front left \& back left  & rightward  & Source \\
                    \midrule
                    back right \& front left \& front right  & forward    & Source \\
                    back right \& front left \& front right  & backward   & Source \\
                    back right \& front left \& front right  & leftward   & Source \\
                    back right \& front left \& front right  & rightward  & Target \\
                    \bottomrule
                \end{tabular}
            \end{minipage}
            \caption{Context configurations for the Ant environment.}
            \label{tab:ant_contexts}
        \end{table*}

    \subsection{Assistive Gym Environment Setting}
    We further evaluate ConTraIRL in Assistive Gym using a Sawyer robot and a simulated human. Each context combines a pillar configuration with one of three assistive goals: feeding, drinking, or scratching an itch.
    
    \paragraph{Goal factors.} Feeding requires the robot to deliver food to the human's mouth, drinking requires it to bring a cup to the mouth, and scratching requires it to reach and scratch a designated itch location.
    
    \paragraph{Dynamics factors.} We randomly sample ten pillar positions around the robot to define ten dynamics factors. Each factor places one pillar at its corresponding position, which remains fixed within that context. The robot must complete the assistive task without colliding with the pillar. Each pillar position therefore imposes different collision constraints and induces different feasible trajectories and state occupancies.
    
    \paragraph{Context partitions and expert data.} Combining the ten dynamics factors with the three goal factors yields 30 possible contexts. Selected combinations serve as source contexts, while the held-out combinations serve as target contexts. Each target context is source-supported: its pillar configuration and assistive goal each appear in other source contexts. All methods use identical context partitions and expert data. Each source context provides full state-only expert trajectories with phase labels. Unless otherwise specified, 20\% of the states from one target expert trajectory are randomly sampled as anchors at the beginning of training and remain fixed. No other target expert states, expert actions, or task rewards are provided to the IRL methods.
    
    \paragraph{Evaluation.} Policies are optimized using the learned rewards and evaluated using the Assistive Gym task rewards. We report normalized return and success rate, where success is determined by the completion criterion of each assistive task.

    \subsection{Hyperparameters}
        We report the hyperparameters used for all experiments. Unless otherwise stated, the same settings are used across environments and across all baselines for a fair comparison. The hyperparameter for ConTraIRL is listed in Table \ref{tab:contrairl_hyperparameters}. Policy optimization is performed using Soft Actor-Critic (SAC) implemented in Stable-Baselines3. Unless otherwise specified, default SB3 settings are used, with the hyperparameters listed in Table~\ref{tab:contrairl_hyperparameters}.
        \begin{table*}[!ht]
        \centering
        \small
        \begin{minipage}{0.45\textwidth}
        \centering  
        \textbf{Encoder \& Contrastive Learning}

        \begin{tabular}{ll}
        \toprule
        Parameter & Value \\
        \midrule
        Dynamics latent dim & 4 \\
        Dynamics hidden dims & [32,32,32,32] \\
        Goal latent dim & 2 \\
        Goal hidden dims & [8,8,8] \\
        Encoder update steps & 50 \\
        Encoder LR & 1e-3 \\
        Encoder weight decay & 1e-4 \\
        State normalization & True \\
        Contrastive batch size & 4800 \\
        Dynamics diff margin & 0.1 \\
        Dynamics latent bandwidth & 0.2 \\
        Goal diff margin & 0.1 \\
        Goal latent bandwidth & 0.2 \\
        Expert buffer update steps & 500 \\
        Temperature & 0.07 \\
        Number of seeds & 25 \\
        \bottomrule
        \end{tabular}
        \end{minipage}
        \hfill
        \begin{minipage}{0.45\textwidth}
        \centering
        \textbf{Policy Optimization (SAC)}

        \begin{tabular}{ll}
        \toprule
        Parameter & Value \\
        \midrule
        Network arch & [256,256] \\
        Policy LR & 1e-3 \\
        Batch size & 1024 \\
        Discount $\gamma$ & 0.99 \\
        Polyak update & 0.01 \\
        State normalization & True \\
        Weight decay & 5e-4 \\
        Action noise & OU, std=0.2 \\
        Gradient steps & 1 \\
        Train freq & 1 \\
        Entropy coef & auto \\
        Target entropy & auto \\
        Learning starts & 1000 \\
        Stats window & 100 \\
        Policy updates/context & 2000 \\
        \bottomrule
        \end{tabular}
        \end{minipage}
        \caption{Hyperparameters of ConTraIRL used in all experiments.}
        \label{tab:contrairl_hyperparameters}
        \end{table*}

    \subsection{Inferring Dynamics Labels}

    Our main experiments assume that each expert trajectory is annotated with its dynamics factor. In practice, expert trajectories may be collected without dynamics annotations even when the set of source dynamics factors and their environments is known. We therefore evaluate whether each trajectory can be associated with its dynamics factor using states collected from the source environments.
    
    Because the dynamics variations in our benchmarks are defined by joint functionality, we use only the joint-related values from each state for label inference. This reduces the influence of goal and phase information. We apply \(k\)-means to the joint-state values from all source expert trajectories, setting the number of clusters to the known number of dynamics factors. Each state receives a cluster label, and each trajectory is assigned to the cluster containing the majority of its states.
    
    The cluster indices do not directly identify the corresponding dynamics factors. We therefore collect probe states from each known dynamics environment using uniformly sampled random actions. A \(k\)-nearest-neighbor classifier is trained on the expert joint-state values using their cluster assignments as labels and then applied to the probe states. We enforce a one-to-one correspondence between dynamics factors and clusters by selecting the assignment that maximizes the total number of matching probe-state predictions. This correspondence assigns a dynamics label to each expert trajectory.
    
    The inferred trajectory labels achieve \(95.6\%\) accuracy relative to the ground-truth dynamics labels, which are used only for evaluation. The result indicates that joint-state occupancies are sufficiently distinct to recover dynamics labels in the evaluated environments.
    
    \subsection{Baselines}
    We compare ConTraIRL with six baselines spanning transferable IRL, contextual adversarial IRL, state-only imitation under dynamics variation, successor-feature matching, and contrastive imitation learning.
    
    \paragraph{TraIRL.}
    TraIRL~\cite{trairl} learns a shared state abstraction across environments and recovers a reward in the abstract space for transfer under dynamics variation. For this comparison, TraIRL receives the joint context \(c=(d,g)\) but does not explicitly separate its dynamics and goal factors.
    
    \paragraph{C-AIRL.}
    C-AIRL is our state-only, context-conditioned adaptation of AIRL-ME~\cite{airl-me}. Its adversarial reward model is conditioned on the joint context through the same context-modulated input mechanism used by ConTraIRL. C-AIRL therefore models dynamics and goal jointly, without enforcing cross-factor invariance or phase consistency.
    
    \paragraph{I2L.}
    I2L~\cite{i2l} addresses state-only imitation under transition-dynamics mismatch by introducing an intermediate policy and decomposing imitation into two alternating optimization problems. We condition its learned components on the joint context, allowing it to use demonstrations collected under different dynamics and goals without explicitly factorizing these sources of variation.
    
    \paragraph{RIME.}
    We use the state-only GAIfO-RIME variant of RIME~\cite{rime}, which matches expert and learner state-transition occupancies across multiple dynamics environments. The policy and discriminators are conditioned on the goal identifier, while the dynamics identifier specifies the corresponding interaction environment. This baseline evaluates whether robustness across dynamics is sufficient for transfer to a held-out dynamics--goal pairing.
    
    \paragraph{SFM.}
    SFM~\cite{sfm} directly aligns expert and learner successor features without learning an explicit reward. We condition its feature and policy networks on the joint context, providing a non-adversarial comparison based on feature expectation matching.
    
    \paragraph{PCIL.}
    PCIL~\cite{PCIL} learns a policy-contrastive representation that separates expert and learner observations and defines the imitation reward through cosine similarity to expert representations. Its encoder and policy receive the joint context, but it does not impose factor-specific invariance or consistency constraints.
    
    \paragraph{Supervision and contextual conditioning.}
    Every method is given access to the same source expert trajectories, fixed target-anchor set, context partitions, factor identifiers, and phase labels. Unless otherwise specified, the target anchors comprise \(20\%\) of the states randomly sampled from one expert trajectory in each target context. No method receives expert actions, additional expert observations, or ground-truth rewards during training. Methods use the available observations according to their respective objectives. In particular, GAIfO-RIME uses only valid consecutive state pairs and does not treat independently sampled target anchors as transitions.
    
    For baselines without a native contextual architecture, the dynamics and goal identifiers are provided jointly as \(c=(d,g)\). Thus, these methods receive the same contextual information as ConTraIRL but are not given its factorized representation.
    
    \paragraph{Optimization and hyperparameter selection.}
    Policy optimization uses the same SAC implementation and policy hyperparameters for all methods, while each baseline retains its method-specific reward or representation objective. We perform grid-based hyperparameter tuning using the same validation protocol and search budget for every baseline. Table~\ref{tab:baseline_hyperparameter} reports the selected settings.

        \begin{table}[!ht]
            \centering
            \small
            \setlength{\tabcolsep}{0.8mm}

                \begin{tabular}{ccccc}
                    \toprule
                               & Ant               & HalfCheetah       & Walker            & Swimmer           \\
                    \midrule
                    TraIRL     & $ 0.57 \pm 0.21 $ & $ 0.55 \pm 0.25 $ & $ 0.51 \pm 0.22 $ & $ 0.50 \pm 0.28 $ \\
                    SFM        & $ 0.88 \pm 0.09 $ & $ 0.90 \pm 0.05 $ & $ 0.89 \pm 0.06 $ & $ 0.90 \pm 0.04 $ \\
                    PCIL       & $ 0.89 \pm 0.03 $ & $ 0.92 \pm 0.02 $ & $ 0.91 \pm 0.04 $ & $ 0.88 \pm 0.07 $ \\
                    C-AIRL     & $ 0.82 \pm 0.11 $ & $ 0.85 \pm 0.09 $ & $ 0.86 \pm 0.07 $ & $ 0.84 \pm 0.05 $ \\                    
                    RIME       & $ 0.76 \pm 0.14 $ & $ 0.88 \pm 0.01 $ & $ 0.85 \pm 0.10 $ & $ 0.84 \pm 0.09 $ \\
                    I2L        & $ 0.52 \pm 0.07 $ & $ 0.63 \pm 0.04 $ & $ 0.65 \pm 0.11 $ & $ 0.89 \pm 0.02 $ \\
                    ConTraIRL  & $ \textbf{0.93} \pm \textbf{0.02} $ & $ \textbf{0.95} \pm \textbf{0.01} $ & $ \textbf{0.96} \pm \textbf{0.01} $ & $ \textbf{0.97} \pm \textbf{0.02} $ \\
                    \midrule   
                    Expert     & $ 3393.50 $       & $ 5657.33 $       & $ 3125.94 $       & $ 55.66 $ \\ 
                    \bottomrule
                \end{tabular}
            \caption{Benchmark results on target contexts. For each env, we evaluate on multiple target contexts. Scores are normalized per environment by expert return and then averaged over all target contexts and random seeds (mean $\pm$ std).}
            \label{tab:full_benchmark_target_avg}
        \end{table}

        \begin{table*}[!ht]
            \centering
            \small
            \begin{minipage}{0.48\textwidth}
            \centering
            \textbf{TraIRL}

            \begin{tabular}{ll}
            \toprule
            Parameter & Value \\
            \midrule
            $\lambda_{\mathcal{D}}$ & 0.1 \\
            $\lambda_{\text{VAE}}$ & 1.0 \\
            $\lambda_{\text{WGAN}}$ & 1.0 \\
            $\lambda_{\mathcal{F}}$ & 1.0 \\
            \midrule
            Reward LR & 3e-4 \\
            Batch size & 2048 \\
            Weight decay & 1e-3 \\
            Reward net & [32,32,32] \\
            Activation & Tanh \\
            Reward updates & 10 \\
            \midrule
            Encoder net & [32,32,32] \\
            Decoder net & [32,32,32] \\
            Latent dim & 8 \\
            VAE LR & 3e-4 \\
            VAE updates & 10 \\
            Disc net & [32,32] \\
            Disc updates & 10 \\
            \bottomrule
            \end{tabular}

            \vspace{3mm}
            
            \textbf{SFM}          
            
            \begin{tabular}{ll}
            \toprule
            Parameter & Value \\
            \midrule
            SF discount & 0.99 \\
            SF LR & 5e-4 \\
            Actor LR & 5e-4 \\
            Feature function & adversarial discriminator \\
            Batch size & 2048 \\
            Feature LR & 5e-4 \\
            Weight decay & 1e-3 \\
            Activation & Tanh \\
            Target noise & 0.2 \\
            Noise clip & 0.5 \\
            Action noise & 0.1 \\
            \bottomrule
            \end{tabular}

            \end{minipage}
            \hfill
            \begin{minipage}{0.48\textwidth}
            \centering
            \textbf{PCIL}
                        
            \begin{tabular}{ll}
            \toprule
            Parameter & Value \\
            \midrule
            Update frequency & 2 \\
            Learning rate & 1e-4 \\
            Batch size & 128 \\
            Hidden dim & 256 \\
            Temperature & 0.07 \\
            \bottomrule
            \end{tabular}

            \vspace{3mm}
            \textbf{RIME}

            \begin{tabular}{ll}
                \toprule
                Parameter & Value \\
                \midrule
                Discriminator net & [100,100] \\
                Batch size & 2048 \\
                Discriminator update epochs & 5 \\
                Gradient penalty coefficient & 10 \\
                Entropy coefficient & 0 \\
                \bottomrule
            \end{tabular}

            \vspace{3mm}
            \textbf{C-AIRL}
            
            \begin{tabular}{ll}
            \toprule
            Parameter & Value \\
            \midrule
            $g_\theta(s)$ net & [64,64,64] \\
            $h_{\phi^i}(s)$ net & [64,64,64] \\
            Learning rate & 3e-4 \\
            Batch size & 2048 \\
            Weight decay & 1e-3 \\
            Activation & Tanh \\
            Disc gradient steps & 10 \\
            \bottomrule
            \end{tabular}

            \vspace{3mm}
            \textbf{I2L}
            
            \begin{tabular}{ll}
            \toprule
            Parameter & Value \\
            \midrule
            Discriminator net & [64,64] \\
            Discriminator LR & 3e-4 \\
            Discriminator update steps & 5 \\
            Wasserstein critic net & [64,64,64] \\
            Wasserstein critic LR & 5e-5 \\
            Wasserstein critic update steps & 20 \\
            Activation & Tanh \\
            Priority buffer capacity & 5 trajectories \\
            Priority buffer refresh rate & 5 \\
            Expert batch size & 500 \\
            \bottomrule
            \end{tabular}
            
            \end{minipage}
            \caption{Hyperparameters for baseline methods.}
            \label{tab:baseline_hyperparameter}
        \end{table*}

\section{Further Discussions} \label{appendix:discussion}

\subsection{Connection to Zero-Shot Compositional Transfer}
ConTraIRL considers compositional transfer with limited target supervision rather than zero-shot transfer. Source support ensures that each target factor has appeared in the source contexts, but it does not ensure that source trajectories cover the states induced by their target pairing. When the factor-conditioned target occupancy differs from the corresponding source occupancies, source demonstrations alone cannot identify how target-specific states should be represented or rewarded. Target anchors address this shift by aligning target states with source-derived expert abstractions for the same factor and phase, thereby calibrating the reward on the target occupancy. If the source and target factor-conditioned occupancies coincide, transfer without anchors may be possible; however, fully zero-shot transfer is outside the scope of this work.

\subsection{Phase Labels}
    A phase label \(\psi\in\Psi=\{\psi_1,\ldots,\psi_K\}\) denotes a discrete behavioral stage that is comparable across trajectories and contexts. For each environment, the phase is assigned by an environment-defined rule that combines multiple observable state and temporal quantities. Depending on the environment, these quantities include joint values, velocities, accelerations, spatial locations of relevant bodies or objects, and normalized elapsed time. The combined information is mapped to \(K=20\) stages using rules fixed before training. Although the stages describe the progression of behavior, they are not reward values, reward estimates, or measures of policy performance.
    
    The phase-assignment rules may differ across dynamics contexts while preserving the semantics of the resulting labels. For example, an agent with three disabled legs may exhibit different joint configurations, velocities, and accelerations from a fully functional agent during the same behavioral stage. Applying identical thresholds to individual state quantities could therefore assign behaviorally corresponding states to different phases. We use dynamics-specific rules so that the same phase label represents the same behavioral stage across contexts. Each rule combines the relevant observable quantities rather than relying on any single variable. The rules are fixed before training and applied identically to expert and learner states within each context. The learning algorithm does not infer these rules, and neither expert actions nor reward values are used to define or assign the phase labels.
    
    Phase labels are used to construct expert pairs, align expert and learner states at corresponding behavioral stages, and index the phase-conditioned expert abstractions \(z_d^E(\psi)\) and \(z_g^E(\psi)\). They are not provided as inputs to the encoders, which remain defined as \(\Phi_d(s,d)\) and \(\Phi_g(s,g)\). Phase conditioning prevents states from different behavioral stages from being aligned solely because they share the same dynamics or goal factor. The current formulation therefore assumes access to environment-defined phase labels whose behavioral semantics are consistent across contexts.

\subsection{Limitations}
ConTraIRL has two main limitations. First, its contrastive objectives require repeatedly sampling expert--expert and expert--learner pairs. Moreover, because the encoders change during training, the phase-conditioned expected expert abstractions must be updated throughout optimization. These operations increase the computational cost as the numbers of contexts, factors, phases, and contrastive samples grow. More efficient pair sampling, memory banks, or running estimates of the expert abstractions could reduce this cost.

Second, ConTraIRL assumes that phase labels are available and comparable across contexts. We do not develop a self-supervised method for inferring phases from trajectories. This assumption may be restrictive for behaviors with variable durations, nonmonotonic progress, or cyclic structure. Learning phase representations jointly with the factorized abstractions is an important direction for future work.

\section{Additional Visualizations}
\begin{figure}[!ht]
            \centering
            \begin{minipage}{0.48\linewidth}
                \centering
                \includegraphics[width=\linewidth]{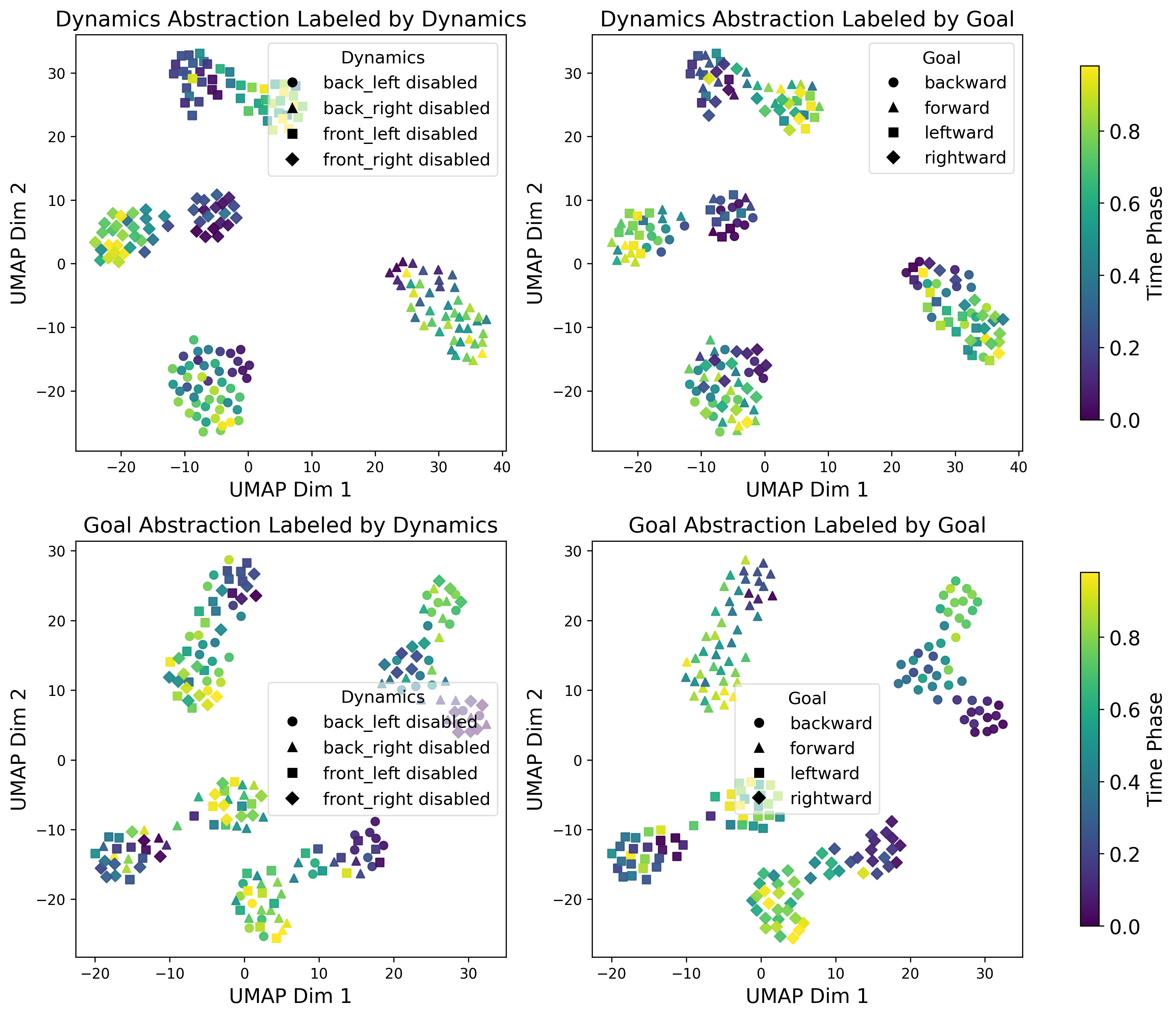}
            \end{minipage}
            \hfill
            \begin{minipage}{0.48\linewidth}
                \centering
                \includegraphics[width=\linewidth]{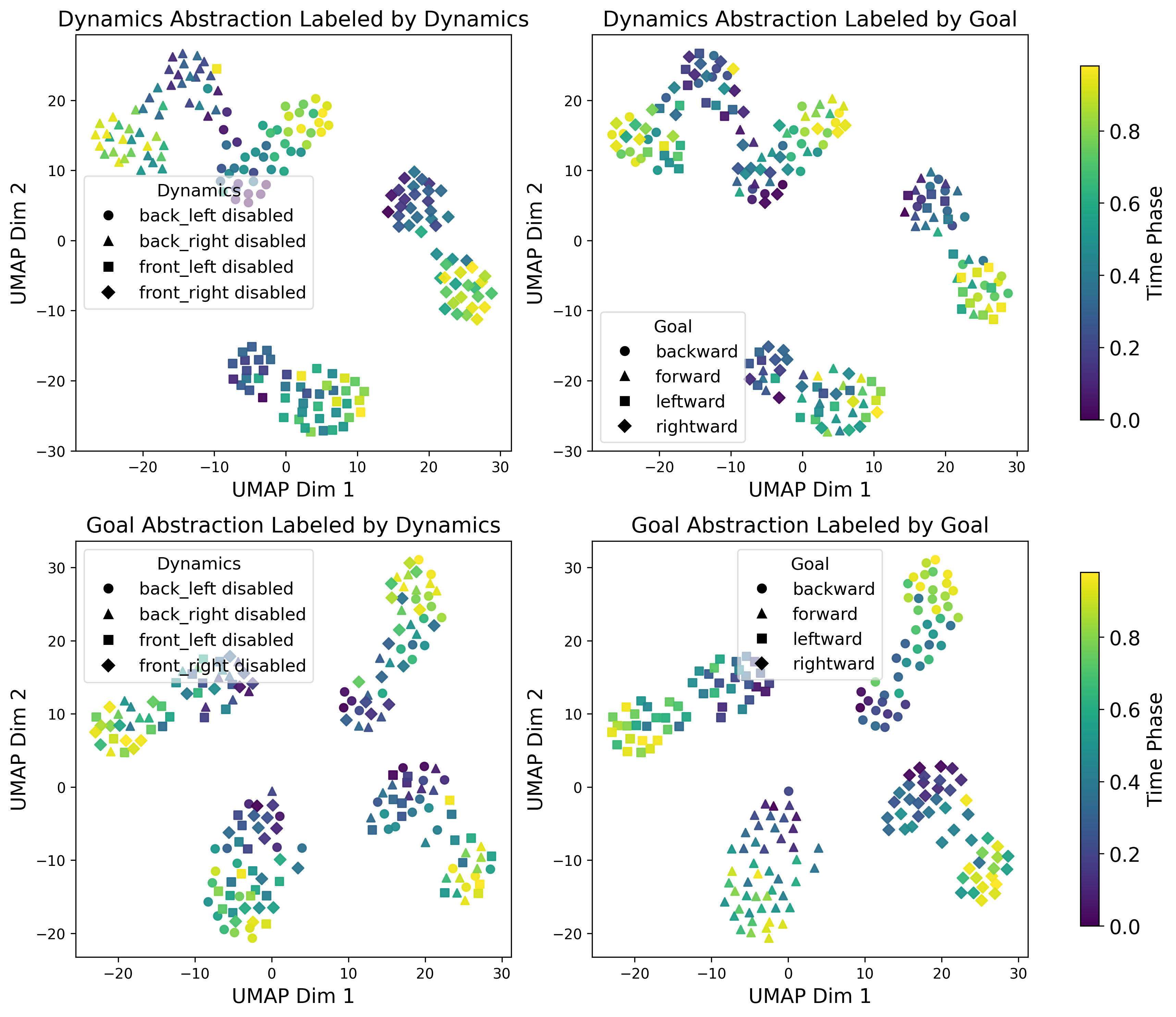}
            \end{minipage}
            \caption{Additional UMAP visualization of the learned abstractions.}
            \label{fig:additional_abstraction_visualization}
        \end{figure}

        Figure \ref{fig:additional_abstraction_visualization} provides additional visualizations of the learned abstractions under different contextual configurations. The dynamics latent space exhibits separation according to dynamics labels while mixing goal labels, whereas the goal latent space shows the complementary pattern. This cross-label mixing confirms factor orthogonality. Within each cluster, color gradients corresponding to normalized temporal phase demonstrate smooth phase consistency in the latent geometry. These results further validate that the learned abstractions capture factor-specific structure while preserving temporal progression.

\begin{figure}[!ht]
            \centering
            \begin{minipage}{0.48\linewidth}
                \centering
                \includegraphics[width=\linewidth]{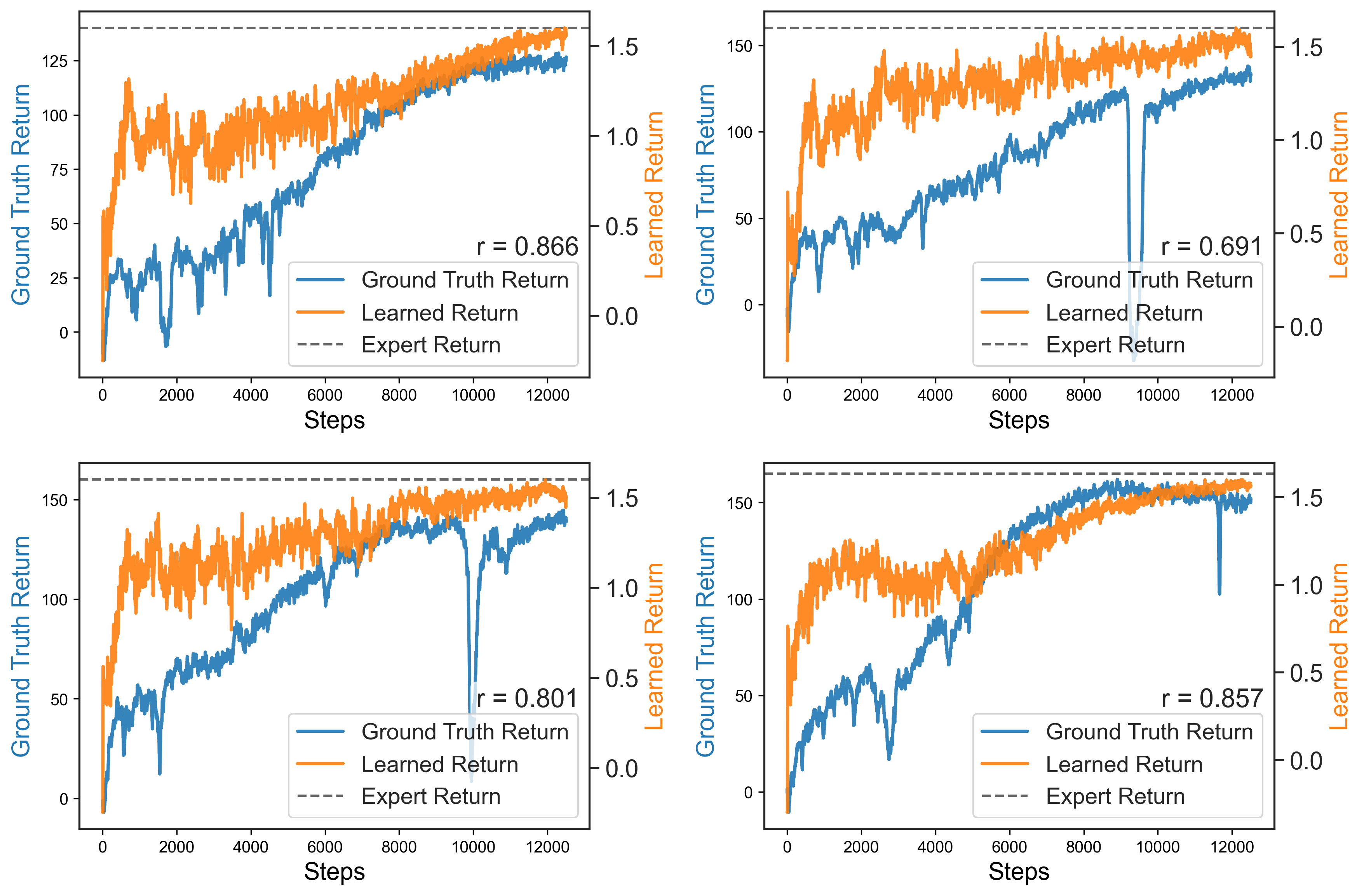}
            \end{minipage}
            \hfill
            \begin{minipage}{0.48\linewidth}
                \centering
                \includegraphics[width=\linewidth]{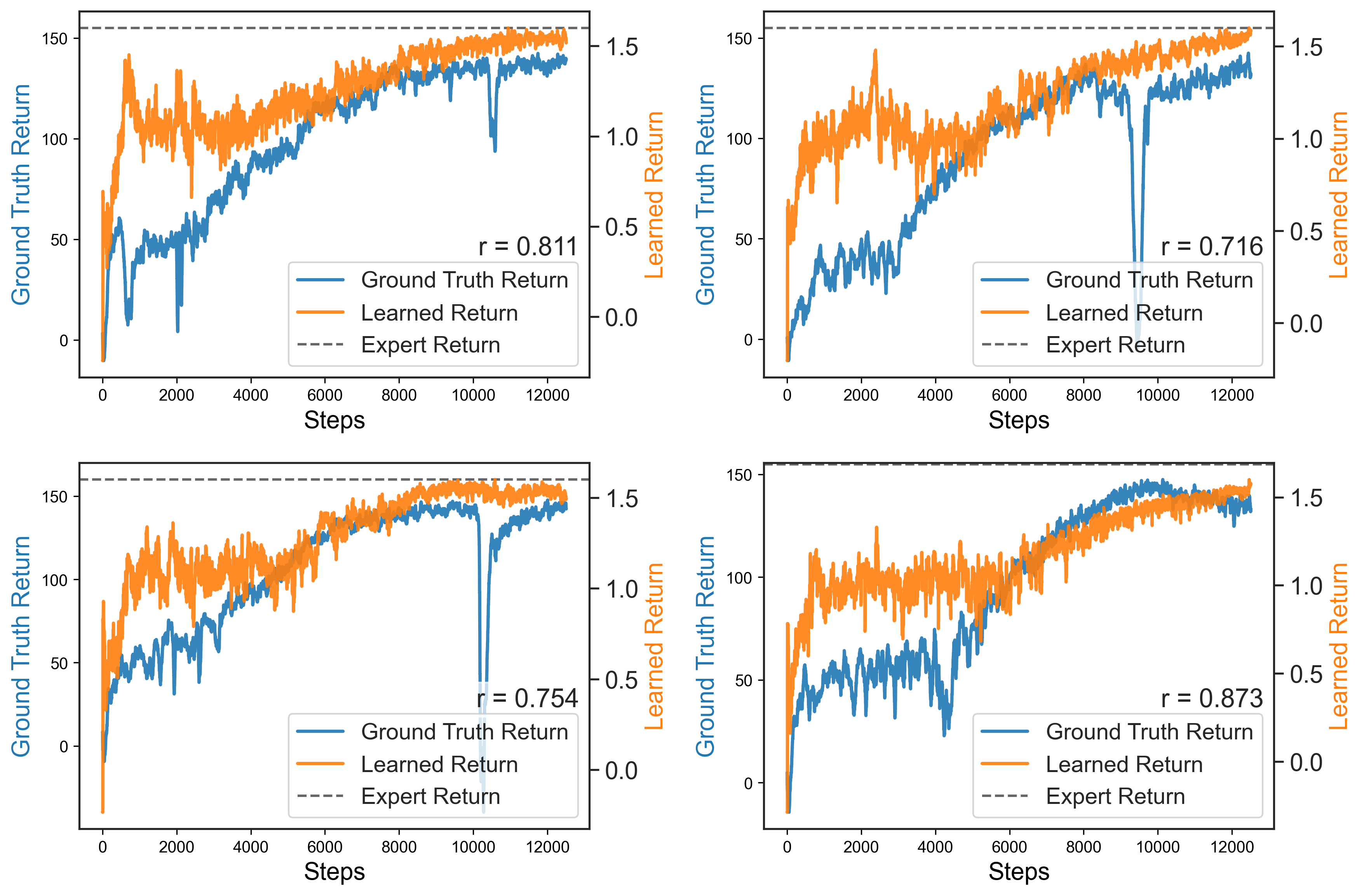}
            \end{minipage}
            \caption{Additional figures of learning curve.}
            \label{fig:additional_learning_curve}
        \end{figure}

        Figure \ref{fig:additional_learning_curve} presents additional learning curves across environments and contextual splits. In all cases, the return computed under the learned reward closely tracks the return under the ground-truth environment reward throughout training. The reported Pearson correlation coefficients ($r$) quantify this alignment and remain consistently high across settings. This behavior indicates that the recovered reward provides a stable and informative objective for policy optimization. The dashed line denotes expert performance for reference.

\section{Computational Resources}
    All experiments were conducted on a single workstation with an Intel Core i7-10700K CPU, an NVIDIA GeForce RTX 3070 GPU, and 32 GB of RAM, running Ubuntu 22.04, CUDA 12.6, python 3.12 and PyTorch 2.10. Table~\ref{tab:training_time} reports the wall-clock training time for one run of each method.
    
    \begin{table}[!ht]
    \centering
    \small
    \begin{tabular}{lc}
        \toprule
        Environment & Training time (hours) \\
        \midrule
        Ant           & 26 hrs \\
        HalfCheetah   & 16 hrs \\
        Walker2d      & 12 hrs \\
        Swimmer       & 8 hrs \\
        Assistive Gym & 38 hrs \\
        \bottomrule
    \end{tabular}
    \caption{Wall-clock training time for one ConTraIRL run.}
    \label{tab:training_time}
    \end{table}

\end{document}